\documentclass[lettersize,journal]{IEEEtran}
\usepackage{amsmath,amsfonts}
\usepackage{algorithmic}
\usepackage{algorithm}
\usepackage{array}
\usepackage{textcomp}
\usepackage{stfloats}
\usepackage{url}
\usepackage{verbatim}
\usepackage{graphicx}
\usepackage{cite}
\DeclareUnicodeCharacter{2212}{-}
\usepackage{xcolor}
\usepackage{booktabs}
\usepackage{multirow}
\usepackage{amssymb}
\usepackage{pifont}
\usepackage{subfigure}

\usepackage{newfloat}
\usepackage{listings}

\begin{document}

\author{
Chunpu Liu \quad Guanglei Yang \quad Wangmeng Zuo \quad Tianyi Zang\\
Harbin Institute of Technology, China  
 \\

}


\title{MetricDepth: Enhancing Monocular Depth Estimation with Deep Metric Learning}




\maketitle

\begin{abstract}
  Deep metric learning aims to learn features relying on the consistency or divergence of class labels. 
  However, in monocular depth estimation, the absence of a natural definition of class poses challenges
  in the leveraging of deep metric learning.
  Addressing this gap, this paper introduces MetricDepth, a novel method that integrates deep metric learning
  to enhance the performance of monocular depth estimation.
  To overcome the inapplicability of the class-based sample identification in previous deep metric learning methods
  to monocular depth estimation task, we design the differential-based sample identification.
  This innovative approach identifies feature samples as different sample types by their depth differentials relative to anchor,
  laying a foundation for feature regularizing in monocular depth estimation models.
  Building upon this advancement, we then address another critical problem caused by the vast range
  and the continuity of depth annotations in monocular depth estimation.
  The extensive and continuous annotations lead to the diverse differentials of negative samples to anchor feature,
  representing the varied impact of negative samples during feature regularizing.
  Recognizing the inadequacy of the uniform strategy in previous deep metric learning methods for handling negative samples
  in monocular depth estimation task, we propose the multi-range strategy.
  Through further distinction on negative samples according to depth differential ranges
  and implementation of diverse regularizing, our multi-range strategy facilitates differentiated regularization
  interactions between anchor feature and its negative samples.
  Experiments across various datasets and model types demonstrate the effectiveness and versatility of MetricDepth,
  confirming its potential for performance enhancement in monocular depth estimation task.
 
\end{abstract}
\section{Introduction}


Depth estimation, a long-lasting research topic in computer vision, seeks to generate corresponding depth maps from given RGB images. 
 This process endows machines with critical depth perception, thereby enhancing their understanding of the real 3D world.
 Its applications are vast and varied, including augmented reality (AR) \cite{azuma1997survey,carmigniani2011augmented,zhang2023controlvideo},
 robotics \cite{siciliano2008springer,karoly2020deep}, and autonomous driving \cite{yurtsever2020survey,levinson2011towards}.
 Within the realm of depth estimation, methodologies can be broadly categorized into several types,
 including stereo-based \cite{hirschmuller2007stereo,geiger2010efficient,chang2018pyramid},
 MVS-based \cite{yao2018mvsnet,gu2020cascade},
 and monocular depth estimation \cite{eigen2014depth,zhao2020monocular,ming2021deep}.
 Of these depth estimation methods, monocular depth estimation (MDE) has garnered increasing interest
 due to its minimal sensor requirement and easy configuration. 
 Predicting depth from a single RGB image is a challenging task,
 as key 3D geometric cues are lost in the projection to 2D.
 Owing to the powerful learning ability of
 deep neural networks \cite{he2016deep,dosovitskiy2020image,liu2021swin}, the availability of manually labeled
 datasets \cite{silberman2012indoor,geiger2013vision}, and the evolution of computation hardware,
 MDE methods have achieved remarkable progress in recent years.

\begin{figure}[!t]
    \centering
    \includegraphics[width=0.7\columnwidth]{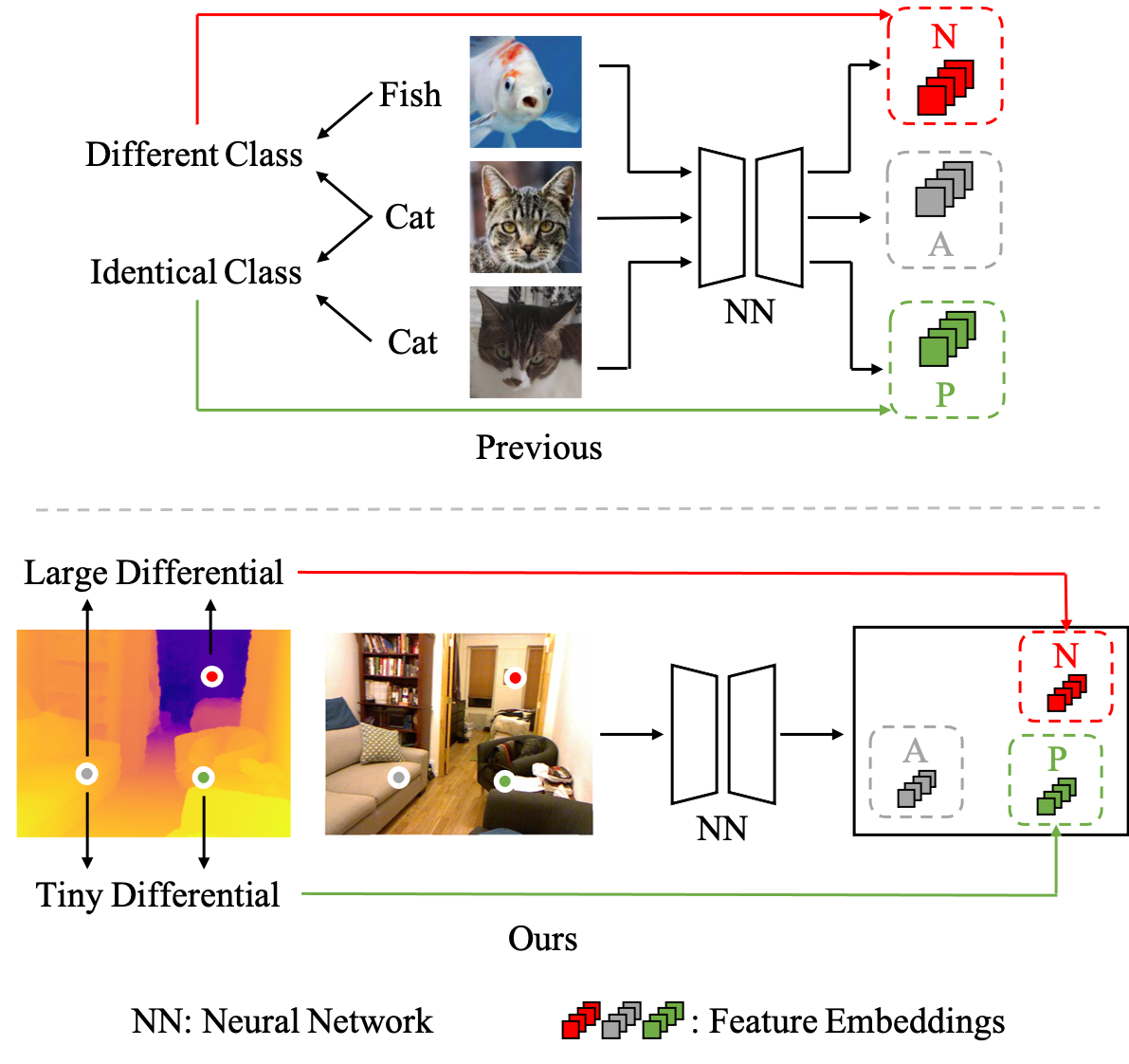}
    \caption{
    An illustration on the main difference between previous deep metric learning methods and our method. 
    \textbf{A} denotes anchor, \textbf{P} denotes positive sample, and \textbf{N} denotes negative sample.
    In previous deep metric learning methods, it relies on class labels to distinguish sample types
    between different features.
    For monocular depth estimation task where no class labels exist, our method relies on depth differentials between features
    to identify sample types.
    }
    \label{fig:teaser}
\end{figure}

Within the spectrum of MDE methods, supervised MDE approaches
 \cite{eigen2014depth,chen2016single,xu2017multi,fu2018deep,lee2019big,bhat2021adabins,yuan2022new,patil2022p3depth}
 stand out for their ability to yield more satisfactory prediction results,
 which is the core focus of our research.
 In conventional supervised MDE methods, deep models are typically trained under pixel-wise constraints on output level,
 aligning the predicted depth maps with depth annotations.
 Beyond employing the straightforward L1 loss, researchers have innovatively tailored various training loss
 terms specific to the MDE task, aiming to enhance the training efficacy of MDE models.
 For instance, Eigen et al. \cite{eigen2014depth} introduced the Scale-Invariant (SI) loss to mitigate the impact
 of depth scale variations,
 and Yin et al. \cite{yin2021virtual} advocated the use of virtual surface normals to augment the robustness of MDE models.
 Despite these advancements in output-level supervision for training MDE models,
 the pursuit of refining feature representations in MDE has not been extensively explored. 
 Recent developments in deep metric learning
 \cite{kaya2019deep,hadsell2006dimensionality,schroff2015facenet,wang2021exploring}
 underscore the pivotal role of high-quality feature representations in augmenting the performance of deep models.
 Deep metric learning (DML) \cite{kaya2019deep} primarily seeks to introduce regularizing in the
 feature space of deep models by referring to the similarities or disparities among features.
 In some areas of computer vision such as image classification and semantic segmentation, a series of DML methods
 \cite{hadsell2006dimensionality,schroff2015facenet,wang2021exploring} have already been extensively investigated.
 Given the demonstrated success of DML in these fields, devising novel DML
 method for MDE task is a promising way to elevate the estimation accuracy of MDE models.


However, the application of DML in MDE task lags significantly behind its advancements in
 image classification and semantic segmentation.
 A primary challenge in adapting DML for MDE task lies in the difficulty of applying the class-based sample identification,
 as used in previous DML methods \cite{hadsell2006dimensionality,schroff2015facenet,wang2021exploring}, to MDE task.
 With previous class-based sample identification, the type of feature sample to anchor
 is determined according to the consistency of class labels with anchor.
 Nevertheless, in MDE task, the ground truth annotations are continuous depth values where no natural definition
 of class exists.
 Under above condition in MDE task, the type of feature sample is unlikely to get determined
 with previously adopted class-based sample identification, resulting in the absence of reliance on feature regularizing.

In this study, we introduce MetricDepth, an innovative DML method to enhance the performance of MDE models
 by enabling effective feature regularizing.
 To facilitate feature sample identification in DML for MDE task, we innovatively design the differential-based
 sample identification.
 Considering that in MDE, diverse feature representations contribute to the variation in estimated depth values,
 we utilize these depth values as proxies for characterizing different features,
 rather than relying on traditional class labels.
 This novel sample identification approach distinguishes features in MDE models as different types to anchor
 by referencing their depth differentials instead of the consistency of class labels.
 Specifically, a feature sample is considered a positive sample when the depth differential relative to
 the anchor feature is minimal.
 Conversely, significant depth differential designates a feature sample as negative sample to the anchor.
 The identification of feature sample type is the prerequisite for feature regularizing in DML.
 With the types of feature samples determined via the differential-based identification,
 the regularizing process can then be strategically implemented.
 As usual, the regularizing involves reducing the feature distance between positive samples and the anchor,
 while enlarging the distance between negative samples and the anchor,
 thereby effectively refining the feature space in deep models.

Furthermore, we introduce the multi-range strategy in MetricDepth to
 more effectively harness the potential of negative samples, thereby enhancing the overall regularizing effect.
 In previous DML methods for image classification and segmentation \cite{hadsell2006dimensionality,schroff2015facenet},
 the uniform strategy for identification case and regularizing manner on negative samples is adopted.
 Under the uniform strategy, feature samples with class labels distinct from the anchor are categorized into
 a single negative group and subjected to a consistent regularization formula.
 However, in MDE task, negative samples not only are distinguished to positive samples and anchor feature
 but also show some difference even within the inner negative group.
 Due to the large extent and the continuity of depth annotations, negative samples can have various depth differentials
 to anchor feature, which suggests that negative samples can exert diverse regularizing effect on anchor feature.
 For MDE task, following the uniform strategy as in previous DML methods \cite{hadsell2006dimensionality,schroff2015facenet}
 obviously limits the regularizing potentials of negative samples.
 Under the proposed multi-range strategy,
 negative samples are further distinguished into more fine-grained subgroups according to different ranges of
 depth differentials between them and anchor.
 Besides, between anchor feature and different subgroups of negative samples,
 the multi-range strategy implements diverse regularizing formulas, achieving the differentiation of regularizing
 on negative samples.
 

To validate the efficacy of our method,
 we conduct extensive experiments on popular MDE datasets NYU Depth V2 \cite{silberman2012indoor} (indoor)
 and KITTI \cite{geiger2013vision} (outdoor) with diverse MDE models constructed with different neural network architectures.
 The experimental results show that with the integration of MetricDepth,
 the performance of selected models all get significantly improved,
 which proves the effectiveness and the versatility of our method.

Our contributions are summarized as follows:

\begin{itemize}
    \item We introduce MetricDepth, a pioneering DML method specifically designed for the MDE task.
    By incorporating regularizing on feature level during model training,
    MetricDepth significantly enhances the estimation performance of MDE models.

    \item 
    Addressing the inadequacy of traditional class-based sample identification for MDE,
    we design the novel differential-based sample identification.
    Based on depth differentials, the differential-based sample identification distinguishes feature samples
    as different types to anchor,
    enabling the identification of feature sample types in MDE tasks where no class labels exist.

    \item To optimize the utilization of negative samples characterized by diverse depth differentials,
    we propose the multi-range strategy.
    By further distinguishing negative samples according to different differential ranges
    and implementing diverse regularizing,
    the proposed multi-range strategy achieves differentiated regularizing between anchor feature and its
    negative samples.

    \item The extensive experiments on different datasets with various MDE models demonstrate the effectiveness and
    versatility of our method.

\end{itemize}
\section{Related Work}

\subsection{Monocular Depth Estimation}

In recent years, the fast development of deep learning and the emergency of manually labeled depth datasets such as
 NYU Depth V2 \cite{silberman2012indoor} and KITTI \cite{geiger2013vision} greatly promote the progress of MDE.
 As an early work,
 Eigen et al. \cite{eigen2014depth} first propose to accomplish MDE task with CNN-based neural network model.
 Some subsequent works \cite{liu2015learning,xu2017multi} take into consideration
 the continuous property of MDE task and combine CRFs with CNNs to process deep features for predicting more
 detailed depth results.
 Inspired by the geometry characteristic of MDE task, in BTS \cite{lee2019big} the local planar guidance is proposed
 to preserve the local consistency of intermediate model features in the upsampling process.
 These MDE methods mentioned above usually take MDE as the regression
 task and design models to directly estimate single-channel depth values. Besides the regression-based MDE methods,
 some researchers \cite{cao2017estimating} explore to reformulate MDE as classification task.
 Instead of directly regressing depth values, these methods discrete the allowed maximum depth into a series of
 depth ranges and design models to predict the probability which represents how likely a certain scene location
 may locate at these divided depth ranges.
 Under the concept of classification in MDE methods,
 ordinal constraint are introduced by some researchers \cite{fu2018deep,cao2019monocular,meng2021cornet}
 into the training of MDE models,
 and some adaptive algorithms \cite{bhat2021adabins,piccinelli2023idisc} appear to better solve the depth range division problem.
 In addition to CNN-based MDE models, many transformer-based MDE models
 \cite{ranftl2021vision,yuan2022new} also achieve impressive performance.
 
 
These methods mentioned above mainly focus on the design of MDE models.
 Rather than design more elaborate model architectures, some researchers in the area of MDE employ newly designed
 loss terms in model training to improve the performance of MDE models.
 Eigen et al. \cite{eigen2014depth} propose the
 Scale-Invariant Error \cite{eigen2014depth} to alleviate the influence of global scale on the average loss.
 Yin et al. \cite{yin2021virtual} design a new loss term called virtual normal loss (VNL) to increase the
 robustness of MDE models.
 Although these loss terms \cite{eigen2014depth,yin2021virtual} lead to some improvement on the training
 of MDE models, most of them remain on output level.
 Relevant research about exploring feature constraints to increase the performance of MDE models
 is still limited.
 As far as we know, current MDE methods involving study on feature-level property of MDE models rely on
 the incorporation of additional tasks.

In MDE methods leveraging multi-task learning, MDE models usually gain extra promotion by training together 
 with segmentation models \cite{zhang2018joint,xu2018pad}.
 Though MDE models can learn better features by involving extra guidance from the model features of other tasks,
 there exist some problems.
 One problem is the requirement for extra annotations.
 Among current popular datasets, the RGB-D training samples cannot find enough matched labels of other tasks,
 which causes inconvenience in the joint training of models for multiple tasks.
 In addition, there exist significant difference between the deep features in MDE models and models for other tasks.
 In some multi-task based MDE methods \cite{jung2021fine,chen2023self}, researchers try to explore the relations
 of intermediate features in MDE models by referencing the consistency of semantic labels. 
 However, the consistency or distinction of semantic labels does not entirely equal to the same condition
 on deep features in MDE models.
 In segmentation models, the features corresponding to the same semantic labels are usually highly integrated and
 the features corresponding to diverse labels show obvious difference.
 While in MDE models, this observation dose not always hold due to the metric property of MDE task.
 In this paper, without relying on any extra annotations, the proposed MetricDepth enables exploration of
 feature characteristic in MDE models.

Most of the MDE methods introduced above are supervised MDE methods.
 Besides the supervised MDE methods, there exist other types of MDE methods.
 For example, to avoid the reliance on large-scale labeled depth datasets, the self-supervised MDE methods
 utilize the multiple view geometry priors to train MDE models by view synthesis on sequence video data
 \cite{zhou2017unsupervised,godard2019digging,poggi2020uncertainty,li2022monoindoor++,feng2023iterdepth,liu2023self}
 or stereo data \cite{garg2016unsupervised,godard2017unsupervised}.
 There are also MDE methods aiming to solve the domain shift problems \cite{vankadari2020unsupervised,wang2021regularizing,zhao2022unsupervised,gasperini2023robust,liang2024delving}.
 However, these methods still fall behind the supervised MDE methods in estimation precision. In this paper, we explore
 how to improve the performance of fully supervised MDE methods.

\subsection{Deep Metric Learning}

 By referring to the similarity or discrepancy of training samples,
 deep metric learning methods enable neural network models to learn more
 consistent or discriminating deep features, achieving higher prediction performance.
 Deep metric learning has been successfully applied in several computer vision tasks such as
 recognition \cite{hadsell2006dimensionality,schroff2015facenet,wen2016discriminative}
 and scene segmentation \cite{wang2021exploring,wang2022semi}.
 In the field of deep metric learning, contrastive loss \cite{hadsell2006dimensionality} and
 triplet loss \cite{schroff2015facenet} are two prevalent loss functions
 instrumental in enhancing the representation of feature spaces.
 Contrastive loss \cite{hadsell2006dimensionality}
 aims to reduce the feature distance between positive sample pairs (i.e., similar samples)
 while increasing the feature distance between negative sample pairs (i.e., dissimilar samples).
 This method achieves its optimization goal by computing the Euclidean distance between pairs of samples
 and adjusting the feature distance based on their similarity.
 On the other hand, triplet loss \cite{schroff2015facenet} further expands on this concept.
 Each triplet comprises three parts: an anchor sample, a positive sample (similar to the anchor),
 and a negative sample (dissimilar to the anchor).
 The objective of triplet loss is to ensure that the feature distance between the anchor and the positive sample
 is less than the feature distance between the anchor and the negative sample, within a certain margin.
 Subsequently, more studies on deep metric learning
 such as center loss \cite{wen2016discriminative}, N-pair loss \cite{sohn2016improved}
 and hard sample mining \cite{shrivastava2016training} are developed to further improve its performance.
 In addition to the application in recognition and classification, deep metric learning also
 gets applied in segmentation. The work by Wang et al. \cite{wang2021exploring} is the pioneer research
 expanding the concept of contrastive loss in segmentation. Later a series of studies utilize
 deep metric learning to improve the performance of segmentation task under semi-supervised \cite{wang2022semi} or
 weakly-supervised \cite{yao2021non} settings.

In contrast to the wide application of deep metric learning in these mentioned fields,
 the exploration of deep metric learning in MDE is limited.
 One crucial reason is that the previously proposed deep metric learning methods all follow the class-based
 sample identification to distinguish sample types and regularize deep features in models.
 However, MDE is a metric-oriented task instead of the category-oriented task where the class-based sample
 identification is inconvenient to cut in.
 In this paper, the differential-based sample identification in our method allows the conducting of
 feature sample identification and enables feature regularizing for MDE models.

\section{Method}

\begin{figure*}[h]
    \centering
    \includegraphics[width=0.8\textwidth]{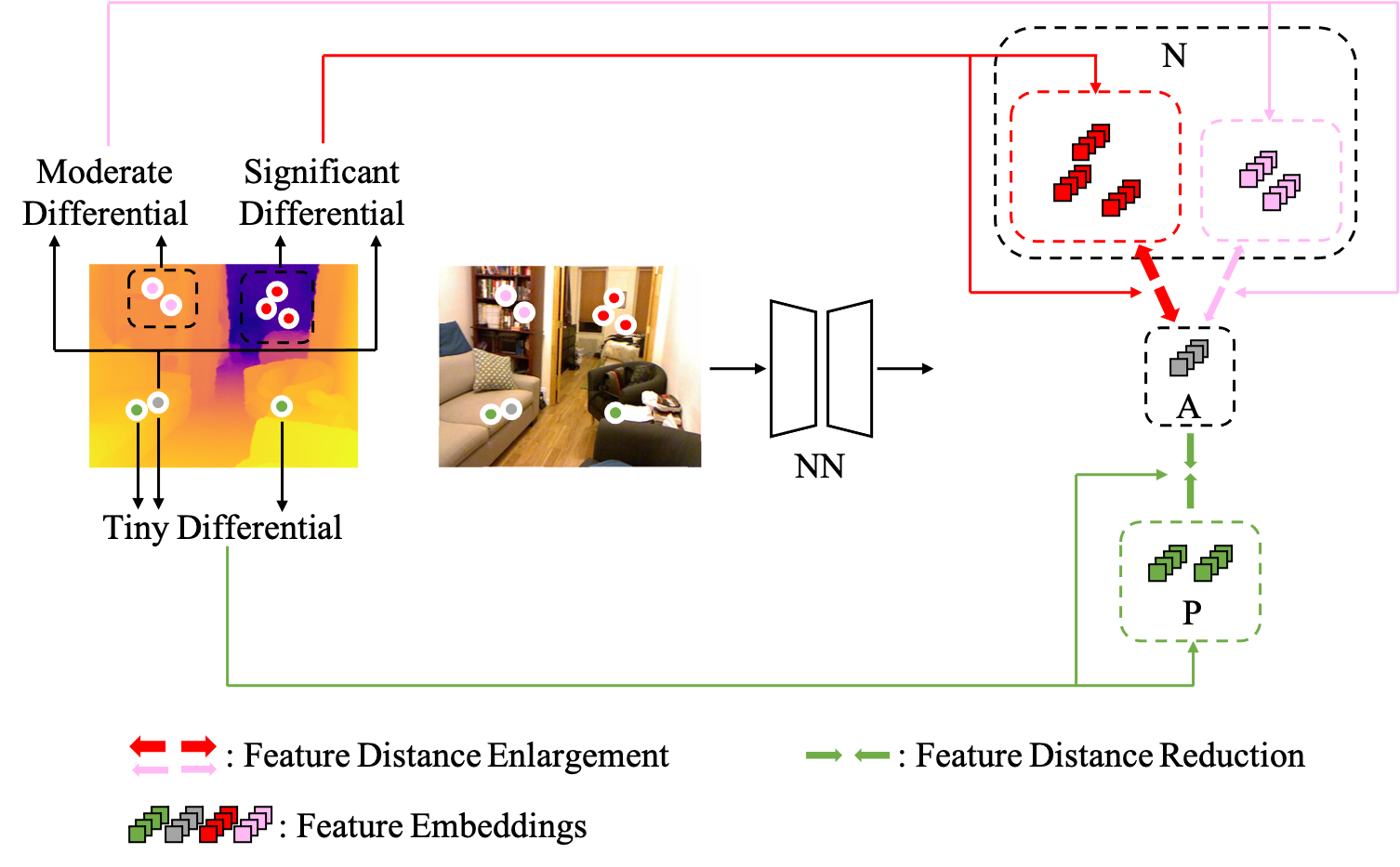}
    \caption{
    An illustration on the core idea of our method.
    \textbf{A} denotes anchor, \textbf{P} denotes positive sample, and \textbf{N} denotes negative sample.
    Feature samples are identified as different types by referencing their depth differentials
    with anchor. For negative samples, the samples with diverse depth differentials are further distinguished
    and implemented with different regularizing formulas.
    }
    \label{fig:pipeline}
\end{figure*}

Given the input RGB image $I$, monocular depth estimation studies how to
 precisely estimate the corresponding depth map $D$ by training model $f$.
 The proposed MetricDepth aims to improve the performance of model $f$ by
 enforcing feature regularizing during the training of $f$.
 In this section, we elaborate our method as follows.
 In Sec. \ref{sec:0}, how feature samples are collected for anchor feature in MetricDepth is firstly introduced.
 Then the designed differential-based sample identification and how it enables feature
 regularizing for MDE models are introduced in Sec. \ref{sec:1}.
 Later the multi-range strategy for better utilizing negative samples is detailed in Sec. \ref{sec:2}.
 At last, the loss functions are presented in Sec. \ref{sec:3}.
 Fig. \ref{fig:pipeline} shows the core idea of our method.

\subsection{Step 0 -- Feature Sample Collecting}
\label{sec:0}

In MetricDepth, deep features in MDE models are regularized by considering the relations of depth differential
 among different scene positions.
 Consequently, a set of features from other scene positions are firstly
 collected for anchor feature as its sample candidates.
 In the process of feature sample collecting, feature samples for anchor feature
 are chosen not only within current feature map, but also from feature maps across the training batch.

For an input RGB image $I$, the to-regularize anchor feature map of $I$ from MDE model $f$ is denoted as $F_{a}$.
 Then the feature samples for $F_{a}$ are collected by:

\subsubsection{Within Current Feature Map}

In this sample collecting means, to efficiently collect feature samples for anchor features
 on all positions in $F_{a}$,
 inspired by the idea of shift window \cite{liu2021swin}, the shift-based approach is adopted.
 For $F_{a} \in \mathbb{R}^{H \times W \times C}$ where $H$ and $W$ denote the height and width of $F_{a}$,
 two random shift seeds $s_{h}$ and $s_{w}$ are firstly generated:
 \begin{equation}
     s_{h}=\Phi(H), s_{h} \in [1,H-1]
 \end{equation}
  \begin{equation}
     s_{w}=\Phi(W), s_{w} \in [1,W-1]
 \end{equation}
where $\Phi(N)$ denotes the function which randomly selects an integer seed from closet $[1,N-1]$.
 Having these two shift seeds, $F_{a}$ is rolled along the vertical direction
 and the horizontal direction respectively with $s_{h}$ and $s_{w}$ to obtain a new feature map $F_{within}$:
 \begin{equation}
     F_{within}=Shift(Shift(F_{a},vertical,s_{h}), horizontal, s_{w})
 \end{equation}
where $Shift(F,dim,n)$ represents the shift operation which rolls the feature map $F$ along $dim$ dimension by $n$ units.
 The features on all positions in $F_{within}$ then  can act as the feature samples for the features in $F_{a}$
 on corresponding positions pixel-wisely.
 By doing the shift sampling as described above, each anchor feature in $F_{a}$ is quickly assigned one
 feature sample at a time.

In addition to shifting feature map, the corresponding depth map needs to
 shift with the same shift seeds $s_{h}$ and $s_{w}$ as well to maintain the locational consistency between
 feature and its depth:
 \begin{equation}
    D_{within}=Shift(Shift(D_{gt},vertical,s_{h}), horizontal, s_{w})
 \end{equation}
where $D_{gt}$ is the ground truth depth annotation paired with $I$.

\begin{figure*}[t]
    \centering
    \subfigure[Within Current Feature Map]{
    \includegraphics[width=0.45\textwidth]{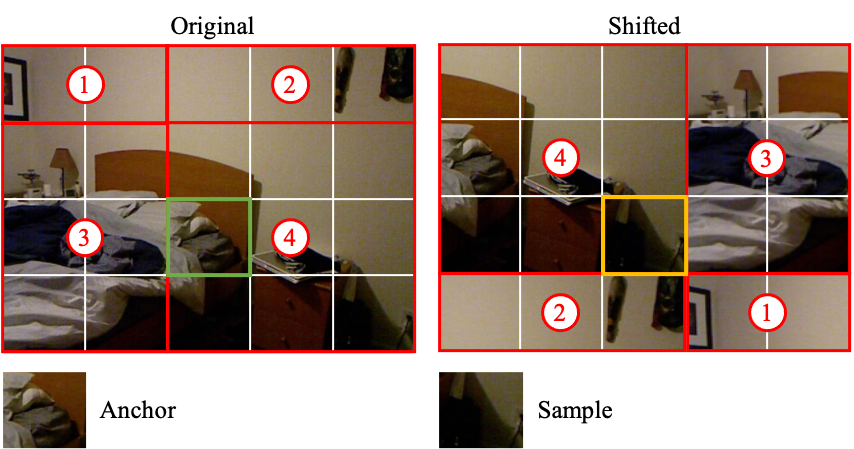}
    }
    \subfigure[Across the Training Batch]{
    \includegraphics[width=0.45\textwidth]{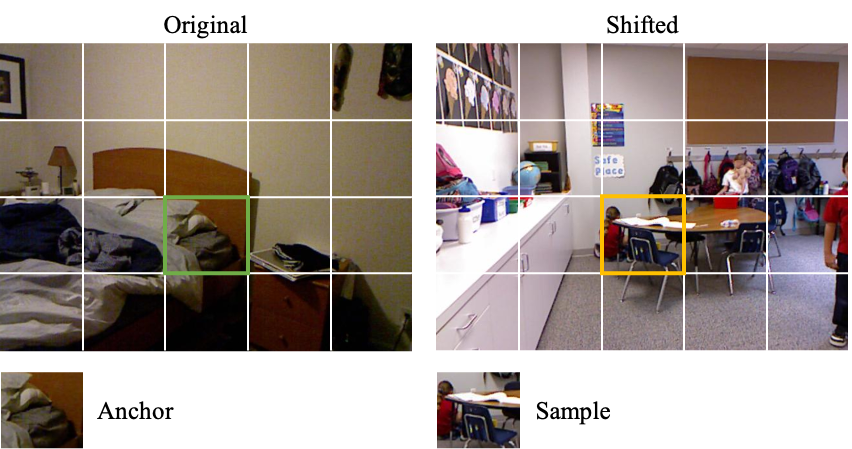}
    }
    \caption{An explanation on the feature sample collecting process in our method.
    In the diagram, the white lines divide the maps into different units.
    In (a), the original map is shifted by 2 units to the left and 1 unit to the top.
    The red boxes and the numbers denote different areas of the map, the green box denotes the anchor on a certain position,
    and the yellow box denotes a sample for the anchor.
    In (b), an entirely different map is shifted to current position
    acting as the sample for the original map.}
    \label{fig:sampling}
\end{figure*}

\subsubsection{Across the Training Batch}

Besides collecting feature samples for anchor features in $F_{a}$ from the same feature map,
 feature samples are also collected from feature maps across the training batch to enrich the
 diversity of samples.
 Assuming that $K$ RGB-D pairs are contained in a single training batch, the feature-depth pairs
 in the batch are denoted as $\{ (F^{0}, D_{gt}^{0}),(F^{1}, D_{gt}^{1}), ..., (F^{K-1}, D_{gt}^{K-1}) \}$.
 A random seed $s_{k}$ is firstly generated as $s_{k}=\Phi(K)$.
 Then the sample feature map $F_{across}$ is chosen as $F^{s_{k}}$ and the shifted depth map $D_{across}$
 is chosen as $D_{gt}^{s_{k}}$.

By iterating these two feature sample collecting means respectively for $N_{within}$ and $N_{across}$ times,
 the set of sample feature maps $S_{F}$ for $F_{a}$ is generated as:
\begin{equation}
\begin{aligned}
    S_{F}= & \{ \overbrace{ F_{within}^{0} , F_{within}^{1} , \cdots ,F_{within}^{N_{within}-1} }^{N_{within}} , \\
    & \overbrace{  F_{across}^{0} , F_{across}^{1} , \cdots ,F_{across}^{N_{across}-1}  }^{N_{across}} \}
\end{aligned}
\end{equation}
And the set of corresponding shifted depth maps $S_{D}$ is generated as:
\begin{equation}
\begin{aligned}
    S_{D}= & \{ \overbrace{ D_{within}^{0} , D_{within}^{1} , \cdots ,D_{within}^{N_{within}-1} }^{N_{within}} , \\
    & \overbrace{  D_{across}^{0} , D_{across}^{1} , \cdots ,D_{across}^{N_{across}-1}  }^{N_{across}} \}
\end{aligned}
\end{equation}

Fig. \ref{fig:sampling} provides an explanation of the feature sample collecting process in MetricDepth.
 In subsequent sections, causing no ambiguity, the feature samples
 in $S_{F}$ and the shifted depth maps in $S_{D}$ are treated equally without distinguishing
 the means by which they are collected.

\subsection{Differential-Based Sample Identification}
\label{sec:1}
 In view of that in MDE, the results derived from differing deep features manifest as diverse depth values,
 the differential of depth values corresponding to different features can serve as the proxy for indicating
 discrepancy of features.
 Intuitively, large depth differentials between two scene positions suggest significant divergence in
 their corresponding features.
 Conversely, small depth differentials indicate a high degree of feature consistency. 
 Out of this motivation, we design the novel differential-based sample identification to complete
 sample type identification for features in MDE models.
 With the differential-based sample identification, the collected feature samples are recognized as different sample
 types according to their depth differentials with respect to anchor feature.

Let us denote the anchor feature map as $F_{a} \in \mathbb{R}^{H \times W \times C}$,
 the depth map of $F_{a}$ as $D_{a} \in \mathbb{R}^{H \times W \times 1}$,
 one sample feature map in $S_{F}$ as $F_{s} \in \mathbb{R}^{H \times W \times C}$,
 the corresponding shifted depth map of $F_{s}$ as $D_{s} \in \mathbb{R}^{H \times W \times 1}$.
 The depth differential map $D_{r}$ is firstly defined as the proxy representing the discrepancy between $F_{a}$ and $F_{s}$.
 The depth differential map $D_{r} \in \mathbb{R}^{H \times W \times 1}$ is calculated as:
\begin{equation}
    D_{r}=|D_{a}-D_{s}|
\end{equation}
Leveraging $D_{r}$, the types of feature samples can be determined by adopting various thresholds
 on depth differentials in $D_{r}$.

As in previous DML methods where feature samples are identified as positive samples and negative samples to anchor,
 two types of differential thresholds are required.
 Assuming that the positive differential threshold is defined as $r_{p}$ and the negative differential
 threshold is defined as $r_{n}$, the identification of feature sample type is conducted as:
\begin{equation}
G_{u}(i)=
\begin{cases}
&0, D_{r}(i) < r_{p}, \\ 
&1, D_{r}(i) > r_{n}
\end{cases}
\label{eq:bi_grouping}
\end{equation}
where $i$ refers to a location in the field of $F_{a}$ and $G_{u} \in \mathbb{R}^{H \times W \times 1}$
 represents the identification map denoting the pixel-wise sample identification results of feature samples
 in $F_{s}$ with respect to $F_{a}$.
 When the depth differential $D_{r}(i)$ is lower than $r_{p}$, $G_{u}(i)$ is set as $0$, which means that
 $F_{s}(i)$ is recognized as the positive sample to $F_{a}(i)$.
 When the depth differential $D_{r}(i)$ is larger than $r_{n}$, $G_{u}(i)$ is set as $1$
 and thus $F_{s}(i)$ is recognized as the negative sample to $F_{a}(i)$.

According to the identification results $G_{u}$, the feature samples belonging to different sample types
 play diverse regularizing effect on anchor feature.
 Following similar idea with the contrastive loss \cite{hadsell2006dimensionality},
 the regularizing loss between the anchor feature and its feature sample on position $i$ is defined as: 
\begin{equation}
    L_{u}(F_{a}(i), F_{s}(i)) = 
    \begin{cases}
    l(F_{a}(i), F_{s}(i)), & \text{if } G_{u}(i) = 0 \\
    \max\left(0, m_{u} - l(F_{a}(i), F_{s}(i))\right), & \text{if } G_{u}(i) = 1
    \end{cases}
    \label{eq:bi_loss}
\end{equation}
where $l(F1,F2)$ denotes the function which calculates the feature distance between $F1$ and $F2$, and $m_{u}$
 refers to the regularizing margin which prevents the over-regularizing between anchor feature and its negative sample.
 When $G_{u}(i)=0$, i.e., $F_{s}(i)$ is a positive sample to $F_{a}(i)$, $L_{u}$ aims to lessen the feature
 distance between $F_{a}(i)$ and $F_{s}(i)$, increasing their consistency.
 When $G_{u}(i)=1$, $L_{u}$ attempts to enlarge the difference in
 feature space between $F_{a}(i)$ and $F_{s}(i)$ since $F_{s}(i)$ plays the role of negative sample for $F_{a}(i)$.
 The total regularizing loss between $F_{a}$ and all $N$ collected feature samples in $S_{F}$ is:
\begin{equation}
L_{re}=\sum_{n=0}^{N-1}  \sum_{i \in \Omega} L_{u}(F_{a}(i),F_{s}^{n}(i))
\end{equation}
where $\Omega$ represents all locations on $F_{a}$.

\subsection{The Multi-Range Strategy on Negative Samples}
\label{sec:2}

\begin{figure}[h]
    \centering
    \subfigure[RGB]{
    \includegraphics[width=0.15\textwidth]{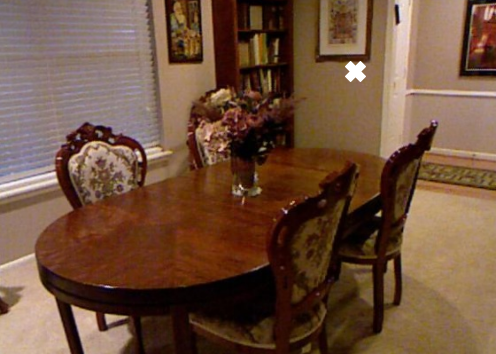}
    }
    \subfigure[Depth Map]{
    \includegraphics[width=0.15\textwidth]{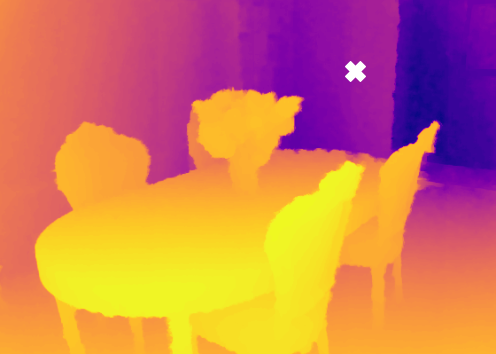}
    }
    \subfigure[Feature Map]{
    \includegraphics[width=0.15\textwidth]{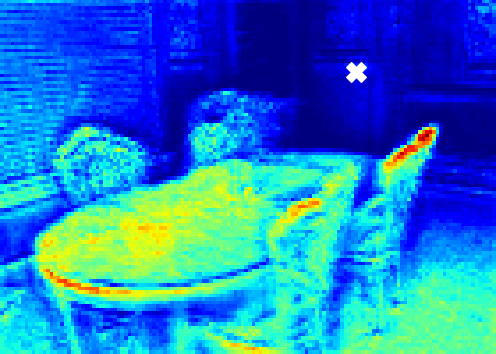}
    }    
    \subfigure[All NS]{
    \includegraphics[width=0.15\textwidth]{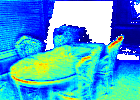}
    }
    \subfigure[Involved NS]{
    \includegraphics[width=0.15\textwidth]{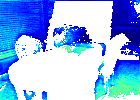}
    }
    \subfigure[Ignored NS]{
    \includegraphics[width=0.15\textwidth]{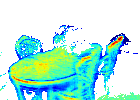}
    }
    \caption{An example exposing the weakness of the uniform strategy on negative samples.
    (a), (b) and (c) show the input RGB image,
    the corresponding depth map, and the feature map of the RGB image.
    The white cross in (c) denotes the anchor feature.
    \textbf{NS} in the captions of (d), (e) and (f) means negative samples.
    In this example, with $r_{n}$ as 0.5, (d) shows all possible negative samples to the anchor feature.
    When $m_{u}$ is set as 2, (e) shows the negative samples which can produce non-zero loss values in Eq. \ref{eq:bi_loss}.
    (f) shows the ignored negative samples which produce no loss value and have no regularizing effect on the anchor feature according to Eq. \ref{eq:bi_loss}.
    }
    \label{fig:ignored_sample}
\end{figure}

As is detailed in Sec. \ref{sec:1}, for MDE task, the identification of positive/negative samples
 and the regularizing between anchor feature and different types of samples can be conveniently
 completed with the designed differential-based sample identification.
 However, the uniform strategy following previous DML methods
 \cite{hadsell2006dimensionality,schroff2015facenet,wang2021exploring}
 for identification case in Eq. \ref{eq:bi_grouping} and relevant regularizing
 formula in Eq. \ref{eq:bi_loss} on negative samples yet achieves sub-optimal performance for MDE task.
 Taking into consideration that the collected negative samples can exhibit diverse depth differentials with anchor feature,
 these negative samples should have various feature distribution, thus resulting in different regularizing impact
 on anchor feature.
 The manner that recognizing these negative samples as a uniform group and applying consistent regularizing
 formula leads to inappropriate regularizing on negative samples.
 As illustrated by the regularization loss in Eq. \ref{eq:bi_loss}, a moderate $m_{u}$ would cause the insufficient
 utilization on the negative samples having large depth differentials with anchor.
 These negative samples inherently should possess significant feature distance with anchor
 since their depth values are greatly varied to anchor.
 Under a moderate margin $m_{u}$, such negative samples are easily filtered out and exert no regularizing
 effect on anchor feature.
 The example in Fig. \ref{fig:ignored_sample} illustrates this case.
 At the same time, setting a large $m_{u}$ could result in the over-regularizing on negative samples
 which have less depth differentials and originally fewer feature distance with anchor.

To more effectively harness the regularization potential of negative samples spanning a vast depth range,
 we propose the multi-range strategy to refine the identification and regularizing on negative samples.
 Instead of identifying feature samples whose depth differentials with anchor feature are larger than $r_{n}$
 as a uniform negative sample group, in the multi-range strategy, the negative feature samples are
 further identified as more fine-grained subgroups according to different discrete ranges of depth differential:
\begin{equation}
G_{ra}(i)=
\begin{cases}
&0, D_{r}(i) < r_{p}, \\ 
&j, r_{l}^{j}<D_{r}(i) < r_{h}^{j}, \\
\end{cases}
\end{equation}
where $j$ denotes the index of the negative sample subgroup, and $r_{l}^{j}$ and $r_{h}^{j}$ denote the lower bound and
 the upper bound of the depth differential range according to which the feature sample $F_{s}(i)$ is classified into the $j$th
 negative subgroup. Similarly, when $G_{ra}(i)=0$, $F_{s}(i)$ acts as the positive sample to $F_{a}(i)$.
 When $G_{ra}(i)\neq0$, $F_{s}(i)$ plays the role of negative sample to $F_{a}(i)$.

In line with the refinement on identification case of negative samples,
 the multi-range strategy advocates implementing layered regularizing formula
 for different subgroups of negative samples:

\begin{equation}
L_{ra}(F_{a}(i), F_{s}(i)) = 
\begin{cases}
l(F_{a}(i), F_{s}(i)), & \text{if } G_{ra}(i) = 0 \\
\max(0, m_{ra}^{j} - l(F_{a}(i), F_{s}(i))), & \text{if } G_{ra}(i) \neq 0
\end{cases}
\label{eq:ra_loss}
\end{equation}

where $m_{ra}^{j}$ denotes the regularizing margin specified for the $j$th negative subgroup.
Under the new identification case and regularizing, the total regularizing loss on all feature samples is:
\begin{equation}
L_{re}=\sum_{n=1}^{N}  \sum_{i \in \Omega} L_{ra}(F_{a}(i),F_{s}^{n}(i))
\end{equation}

By differentiating the identification case and applying adaptive regularizing formula on negative samples, 
 the proposed multi-range strategy achieves better utilization on negative samples,
 thereby enhancing the overall performance of MetricDepth.

\subsection{Loss Functions}
\label{sec:3}

Employing either feature regularizing or output-level constraints in isolation leads to inferior supervision performance in
 the training of MDE models.
 In practice, the combination of regularizing loss provided by MetricDepth and the pixel-wise depth loss $L_{depth}$
 is incorporated to guarantee comprehensive supervision on both feature level and output level.
 In summary, the final loss is:
\begin{equation}
    L_{final}=L_{re}+L_{depth}
    \label{eq:final_loss}
\end{equation}

In Eq. \ref{eq:final_loss}, SI loss \cite{eigen2014depth} is adopted as $L_{depth}$.
 In $L_{re}$, Euclidean distance is adopted as the distance function $l$ for calculating feature distance.

\section{Experiment}


To thoroughly investigate the impact of different components of our method on the prediction performance and
 to validate the overall effectiveness of our method, we have conducted a comprehensive series of experiments.
 This section presents the results of these experiments in detail.
 Firstly, we provide an introduction to the datasets used for these experiments,
 followed by a description of the metrics for performance evaluation in MDE task,
 and the implementation details of our method.
 Subsequently, we report the evaluation results of our method when applied to several MDE models
 built with diverse network architectures.
 Finally, we present our ablation studies, wherein we analyze the influence of individual
 components of our method on its performance.

\subsection{Datasets}

\subsubsection{NYU Depth V2}
The NYU Depth V2 dataset \cite{silberman2012indoor} is collected with the Microsoft Kinect sensor,
 capturing RGBD data across 27 diverse indoor scene categories.
 The collection spans a comprehensive set of 464 distinct indoor scenes such as office and living room, featuring over 400,000 images.
 Notably, each image maintains a resolution of 480×640 pixels, providing rich and varied data for training and evaluating models
 in tasks such as depth estimation, scene understanding, and related computer vision applications. In this work, we follow the data
 split proposed by Eigen et al. \cite{eigen2014depth} for training and evaluating MDE models.

\subsubsection{KITTI}
The KITTI dataset \cite{geiger2013vision}, originally designed for testing computer vision algorithms
 in the context of autonomous driving, includes depth images captured using a LiDAR sensor mounted on a driving vehicle.
 As an outdoor dataset, it provides more than 90k high-resolution pairs of raw data collected in various types of scenes.
 In this work, we also follow the same data split proposed by Eigen et al. \cite{eigen2014depth}
 for model training and evaluation.

\subsection{Evaluation Metrics}
In this work, we follow the same evaluation protocol adopted in previous MDE methods \cite{eigen2014depth,bhat2021adabins,yuan2022new} to
 test the performance of the proposed method. Specifically, the definition of these evaluation metrics are as:
\begin{itemize}
    \item Absolute Relative Difference ($AbsRel$):
    \begin{equation}
        \nonumber
        AbsRel=\frac{1}{N}\sum_{i=1}^{N}\frac{|D_{pred}(i)-D_{gt}(i)|}{D_{gt}(i)}
    \end{equation}
    
    \item Squared Relative Difference ($SqRel$):
    \begin{equation}
        \nonumber
    SqRel=\frac{1}{N} \sum_{i=1}^{N} \frac{|D_{pred}(i)-D_{gt}(i)|^2}{D_{gt}(i)}
    \end{equation}
    
    \item Root Mean Square Error ($RMSE$):
    \begin{equation}
        \nonumber
        RMSE= \sqrt{\frac{1}{N} \sum_{i=1}^{N} |D_{pred}(i)-D_{gt}(i)|^2}
    \end{equation}

    \item Log Root Mean Square Error ($RMSE_{log}$):
    \begin{equation}
        \nonumber
        RMSE_{log}= \sqrt{\frac{1}{N} \sum_{i=1}^{N} |log(D_{pred}(i))-log(D_{gt}(i))|^2}
    \end{equation}
    
    \item Log Average Error ($log10$):
    \begin{equation}
        \nonumber
        log10= \frac{1}{N} \sum_{i=1}^{N} |log_{10}(D_{pred}(i))-log_{10}(D_{gt}(i))| 
    \end{equation}

    \item Threshold Accuracy ($\delta$):
    \begin{equation}
    \nonumber
        \delta _{j}=max\{\frac{D_{pred}(i)}{D_{gt}(i)} , \frac{D_{gt}(i)}{D_{pred}(i)} \}<thr^{j}
    \end{equation}
    where $thr=1.25$ and $j=1,2,3$
\end{itemize}

In above equations, $N$ denotes the total number of pixels, $D_{pred}$ denotes the predicted depth map,
 and $D_{gt}$ represents the ground truth label.
 Among all the evaluation metrics, for $AbsRel$, $SqRel$, $RMSE$, $RMSElog$ and $log10$, lower value represents
 better accuracy.
 For $\delta$, higher value means better performance. In later sections, the evaluation metrics where
 lower value means better performance are labeled with down arrow and other evaluation indexes are labeled with up arrow.

\subsection{Implementation Details}
To show the effectiveness of proposed method, MetricDepth is tested with several representative MDE models
 based on diverse network architectures on both indoor dataset NYU Depth V2 \cite{silberman2012indoor}
 and outdoor dataset KITTI \cite{geiger2013vision}.
 These MDE models are the lightweight UNet model (LT), the CNN-based model BTS \cite{lee2019big},
 and the transformer-based model NeWCRFs \cite{yuan2022new}.
 In the training of all selected MDE models,
 MetricDepth is applied on the feature map before the final prediction layer to generate the regularizing loss.
 In the experiments on BTS and NeWCRFs, we follow the same training configurations
 as claimed in their papers to train models.
 For time efficiency, all the ablation studies are conducted with the LT model on NYU Depth V2 dataset.
 In specific, the LT model takes the MobileNet Large V3 \cite{howard2019searching} as the backbone.
 The decoder of LT is built with a series of layers consisting of stacked upsampling interpolation, 2D convolution,
 and Leaky ReLU activation.
 Additional skip connections between encoder features and decoder features are employed, and one prediction layer is placed
 at the bottom of LT model directly outputting the estimated depth map.
 All experimental results in ablation studies are obtained with the same
 training configuration in NeWCRFs.
 The sample count $N_{within}$ and $N_{across}$ are set as 10 and 4 respectively.
 Relevant experiments about $N_{within}$ and $N_{across}$ are discussed in Sec. \ref{sec:exp_ab} more detailed.
 In practical experiments, setting $r_{p}$ as 0.1 is the suitable choice leading to satisfied performance.
 For different models and different datasets, the negative differential threshold and regularizing margin vary,
 which are discussed in later section and announced in corresponding experiments.
 Our method is implemented with Pytorch \cite{paszke2019pytorch} deep learning library and all experiments are conducted
 with NVIDIA RTX 3090 GPUs.

 \begin{table*}[h]
    \caption{Benchmark performance of selected MDE models on NYU Depth V2.}
    \centering
    \scalebox{0.9}{
    \begin{tabular}{cccccccccc}
    \toprule
      Method &$r_{p}$ & $r_{l}$- $r_{h}$ & $m_{ra}$ & $AbsRel\downarrow$  & $RMSE\downarrow$  & $log10\downarrow$
      & $\delta < 1.25\uparrow $ & $\delta < 1.25^{2}\uparrow$  & $\delta < 1.25^{3}\uparrow $ \\
      \midrule
      
      Eigen et al. \cite{eigen2015predicting}   &- & - &-                      & 0.158  & 0.641 & 0.214    & 0.769 & 0.950  &0.988 \\
      Xu et al. \cite{xu2017multi}              &- & - &-                           & 0.121 & 0.586  & 0.052 & 0.811 & 0.954  &0.987 \\
      DORN \cite{fu2018deep}                    &- & - &-                           & 0.115 & 0.509  & 0.051 & 0.828 & 0.965  &0.992 \\
      VNL \cite{yin2021virtual}                 &- & - &-                           & 0.108   & 0.416  & 0.048 & 0.875 & 0.976  &0.994 \\
      AdaBins \cite{bhat2021adabins}            &- & - &-                           & 0.103   & 0.364 & 0.044 & 0.903 & 0.984  &0.997 \\
      DPT \cite{ranftl2021vision}               &- & - &-                           & 0.110  & 0.357  & 0.045 & 0.904 & 0.988  &0.998 \\
      Zhang et al. \cite{zhang2023improving}       &- & - &-                       & 0.089  & 0.321  & - & 0.932 & -  &- \\
      P3Depth \cite{patil2022p3depth}   &- & - &-                            & 0.104  & 0.356  & 0.043 & 0.898 & 0.981  &0.996 \\
      PixelFormer \cite{agarwal2023attention}   &- & - &-                            & 0.090  & 0.322  & 0.039 & 0.929 & 0.991  &0.998 \\
      URCDC-Depth \cite{shao2023urcdc}           &- & - &-                        & 0.088  & 0.316  & 0.038 & 0.933 & 0.992  &0.998 \\
      
      \midrule
      LT        &- & - &-                           & 0.140  & 0.486 & 0.060 & 0.805 & 0.965  &0.993 \\
      LT+Ours   &0.1 & 0.5-1, 1-1.5, 1.5-2 & 3, 6, 8    & 0.136  & 0.473 & 0.058 & 0.818 & 0.967  &0.994 \\
      \midrule
      BTS \cite{lee2019big} &- & - &-               & 0.119  & 0.419  & 0.051 & 0.865 & 0.975  &0.993 \\
      BTS+Ours  &0.1 & 0.5-1, 1-1.5, 1.5-2 & 2, 4, 6    & 0.115  & 0.405  & 0.049 & 0.871 & 0.979  &0.995 \\
      \midrule
      NeWCRFs \cite{yuan2022new}   &- & - &-                           & 0.089  & 0.321  & 0.038 & 0.930 & 0.991  &0.998 \\
      NeWCRFs+Ours & 0.1 & 0.5-1, 1-1.5 & 5, 9       & 0.087  & 0.317  & 0.037 & 0.934 & 0.992  &0.998 \\
      \bottomrule
    \end{tabular}
    }
    \label{tab:benchmark_nyu}
\end{table*}


\subsection{Benchmark Performance}

In this work, the performance of MetricDepth is evaluated with different MDE models built with diverse network architectures.
 These selected models include the lightweight model (MobileNet V3 Large \cite{howard2019searching}),
 the CNN-based model BTS (ResNet50) \cite{lee2019big},
 and the transformer-based model NeWCRFs (Swin-Large) \cite{yuan2022new}.
 In this section, the evaluation results of these selected models on NYU Depth V2 and KITTI datasets
 after the integration of our method are reported.

\subsubsection{NYU Depth V2}

The evaluation results on NYU Depth V2 dataset of selected models as well as some well-known
 MDE models are shown in Table \ref{tab:benchmark_nyu}.
 The details about the configuration on depth differential threshold and
 regularizing margin for different models are also demonstrated in Table \ref{tab:benchmark_nyu}.
 From the evaluation results in Table \ref{tab:benchmark_nyu} it can be seen that
 the prediction accuracy of selected MDE models all get improved significantly in different degrees
 with the integration of MetricDepth.
 Furthermore, it is noteworthy that promoted by our method, the performance of early published model NeWCRFs \cite{yuan2022new} can surpass or show comparable performance to some recent proposed MDE methods \cite{patil2022p3depth,agarwal2023attention,shao2023urcdc}.

\begin{table*}[t]
    \caption{Benchmark performance of selected MDE models on KITTI.}
    \centering
    \scalebox{0.82}{
    \begin{tabular}{ccccccccccc}
    \toprule
      Method &$r_{p}$ & $r_{l}$- $r_{h}$ & $m_{ra}$ & $AbsRel\downarrow$  & $SqRel\downarrow$  & $RMSE\downarrow$  & $RMSE_{log}\downarrow$
      & $\delta < 1.25\uparrow $ & $\delta < 1.25^{2}\uparrow$  & $\delta < 1.25^{3}\uparrow $ \\
      \midrule
      
      Eigen et al. \cite{eigen2015predicting}   &- & - &-                      & 0.190  & 1.515   & 7.156 & 0.270  & 0.692  & 0.899  &0.967 \\
      DORN \cite{fu2018deep}                    &- & - &-                       & 0.072 & 0.307   & 2.727 & 0.120  & 0.932  & 0.984  &0.995 \\
      VNL \cite{yin2021virtual}                 &- & - &-                       & 0.072 & -       & 3.258 & 0.117  & 0.938  & 0.990  &0.998 \\
      PWA \cite{lee2021patch}                   &- & - &-                       & 0.060 & 0.221   & 2.604 & 0.093  & 0.958  & 0.994  &0.999 \\
      AdaBins \cite{bhat2021adabins}            &- & - &-                       & 0.058 & 0.190   & 2.360 & 0.088  & 0.964  & 0.995  &0.999 \\
      DPT \cite{ranftl2021vision}               &- & - &-                       & 0.062 & -       & 2.573 & 0.092  & 0.959  & 0.995  &0.999 \\
      P3Depth \cite{patil2022p3depth}           &- & - &-                       & 0.071 & 0.270   & 2.842 & 0.103  & 0.953  & 0.993  &0.998 \\
      PixelFormer \cite{agarwal2023attention}   &- & - &-                       & 0.051 & 0.149   & 2.081 & -     & 0.976  & 0.997  &0.998 \\
      URCDC-Depth \cite{shao2023urcdc}          &- & - &-                      & 0.050 & 0.142   & 2.032 & 0.076  & 0.977  & 0.997  &0.999 \\
      \midrule
      LT                                        &- & - &-                       & 0.071  & 0.261 & 2.716 & 0.106 & 0.943 & 0.992  &0.998 \\
      LT+Ours                           &0.1 & 1-5, 5-10, 10-15 & 2, 3, 3.5         & 0.068  & 0.254 & 2.669 & 0.104 & 0.947 & 0.992  &0.998 \\
      \midrule
      BTS \cite{lee2019big}                     &- & - &-                       & 0.061  & 0.250  & 2.803 & 0.098 & 0.954 & 0.992  &0.998 \\
      BTS+Ours                          &0.1 & 1-5, 5-10 & 1, 1.5                 & 0.060  & 0.244  & 2.762 & 0.096 & 0.955 & 0.993  &0.998 \\
      \midrule
      NeWCRFs \cite{yuan2022new}                &- & - &-                       & 0.052  & 0.155  & 2.129 & 0.079 & 0.974 & 0.997  &0.999 \\
      NeWCRFs+Ours                     & 0.1 & 1-5, 5-10 & 4, 7                  & 0.050  & 0.143  & 2.049 & 0.077 & 0.975 & 0.997  &0.999 \\
      \bottomrule
    \end{tabular}
    }
    \label{tab:benchmark_kitti}
\end{table*}

\subsubsection{KITTI}
To validate the efficacy of MetricDepth in outdoor environments, experiments are also conducted on the KITTI dataset
 with the selected models.
 The evaluation results and configuration for MetricDepth on KITTI are shown in Table \ref{tab:benchmark_kitti}.
 Seen from the evaluation results on KITTI dataset, all the evaluation metrics of selected MDE models
 are also improved with the aid of our method,
 confirming the adaptability and effectiveness of our method in outdoor scenarios

\begin{figure}[t]
\centering
    
    \rotatebox[origin=c]{90}{\scriptsize{LT}}
    \subfigure{
        \begin{minipage}{0.16\linewidth}
        \centering
        \includegraphics[height=0.75in,width=1in]{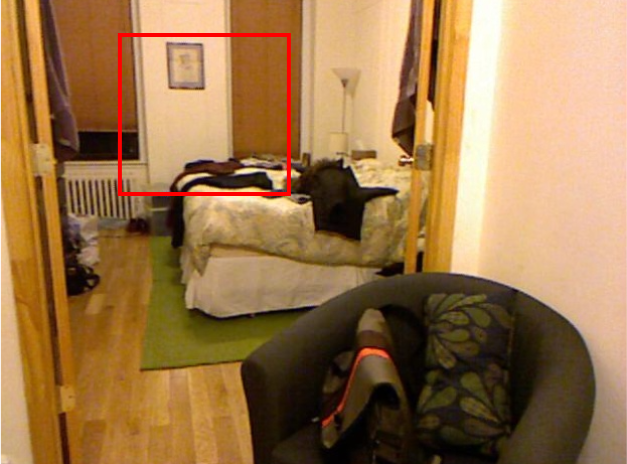}
        \end{minipage}
    }
    \subfigure{
        \begin{minipage}{0.16\linewidth}
        \centering
        \includegraphics[height=0.75in,width=1in]{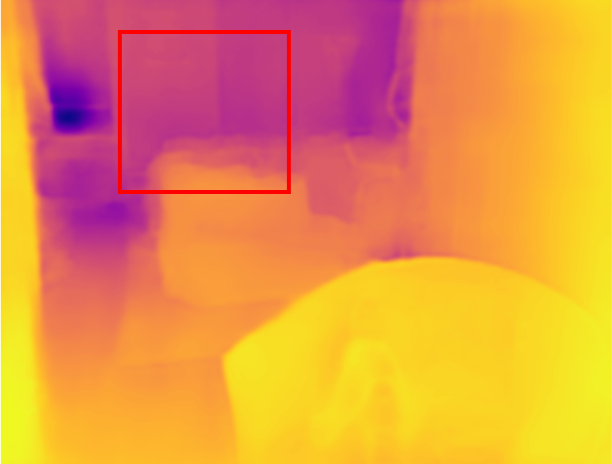}
        \end{minipage}
    }
    \subfigure{
        \begin{minipage}{0.16\linewidth}
        \centering
        \includegraphics[height=0.75in,width=1in]{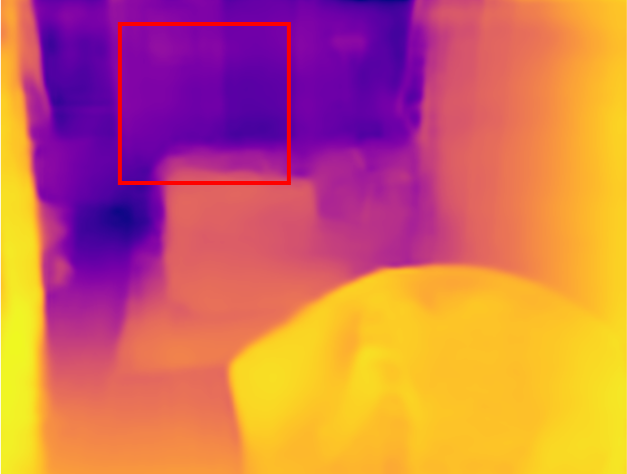}
        \end{minipage}
    }
    \subfigure{
        \begin{minipage}{0.16\linewidth}
        \centering
        \includegraphics[height=0.75in,width=1in]{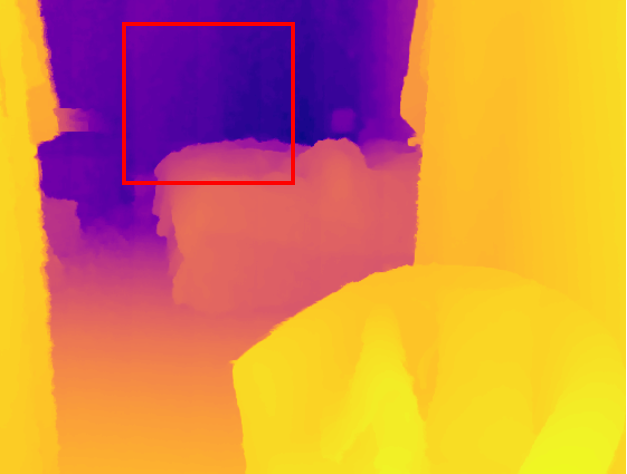}
        \end{minipage}
    }
    
    \vspace{-0.1in}    
    \rotatebox[origin=c]{90}{\scriptsize{BTS}}
    \subfigure{
        \begin{minipage}{0.16\linewidth}
        \centering
        \includegraphics[height=0.75in,width=1in]{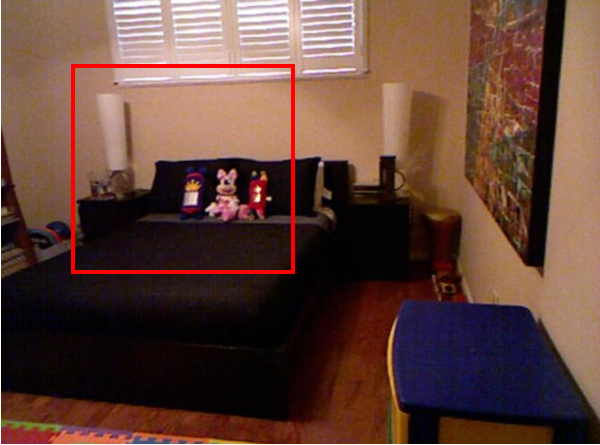}\\
        \end{minipage}
    }
    \subfigure{
        \begin{minipage}{0.16\linewidth}
        \centering
        \includegraphics[height=0.75in,width=1in]{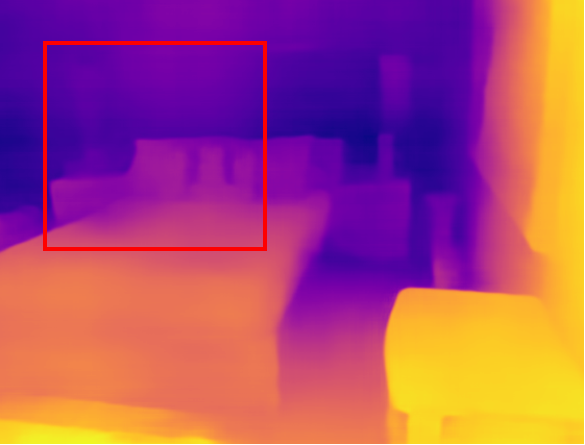}\\
        \end{minipage}
    }
    \subfigure{
        \begin{minipage}{0.16\linewidth}
        \centering
        \includegraphics[height=0.75in,width=1in]{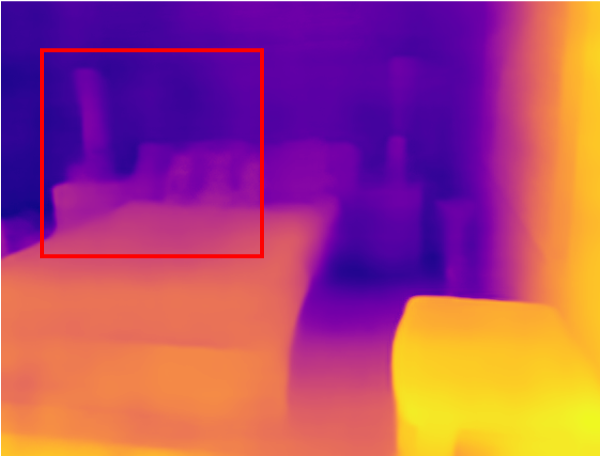}\\
        \end{minipage}
    }
    \subfigure{
        \begin{minipage}{0.16\linewidth}
        \centering
        \includegraphics[height=0.75in,width=1in]{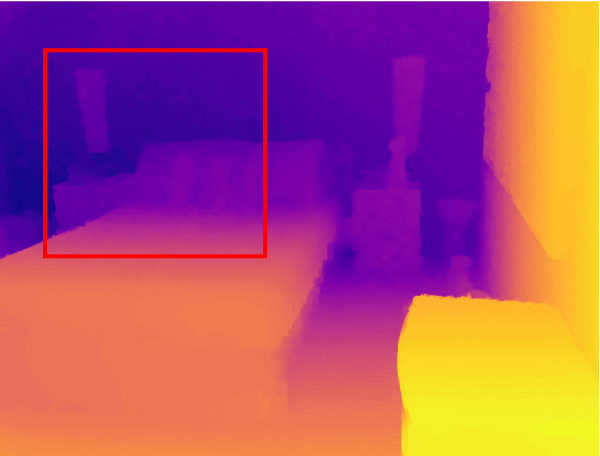}\\
        \end{minipage}
    }
    
    \vspace{-0.1in}
    \setcounter{subfigure}{0}
    
    \rotatebox[origin=c]{90}{\scriptsize{NeWCRFs}}
    \subfigure[RGB]{
        \begin{minipage}{0.16\linewidth}
        \centering
        \includegraphics[height=0.75in,width=1in]{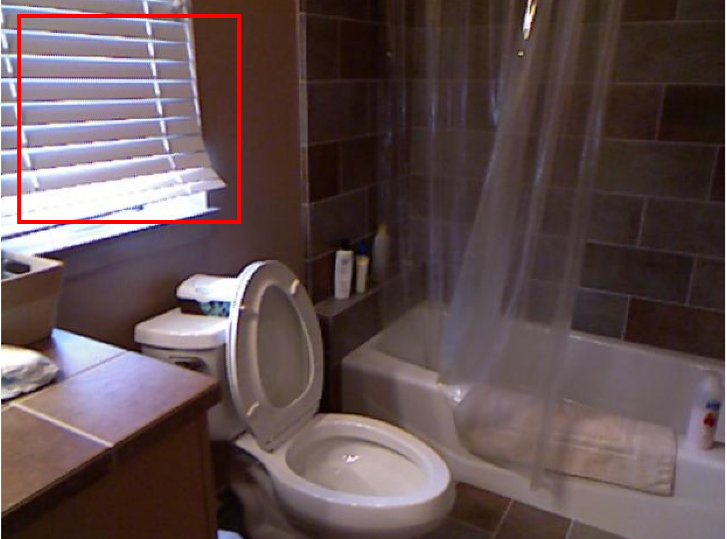} \\
        \end{minipage}
    }
    \subfigure[Original]{
        \begin{minipage}{0.16\linewidth}
        \centering
        \includegraphics[height=0.75in,width=1in]{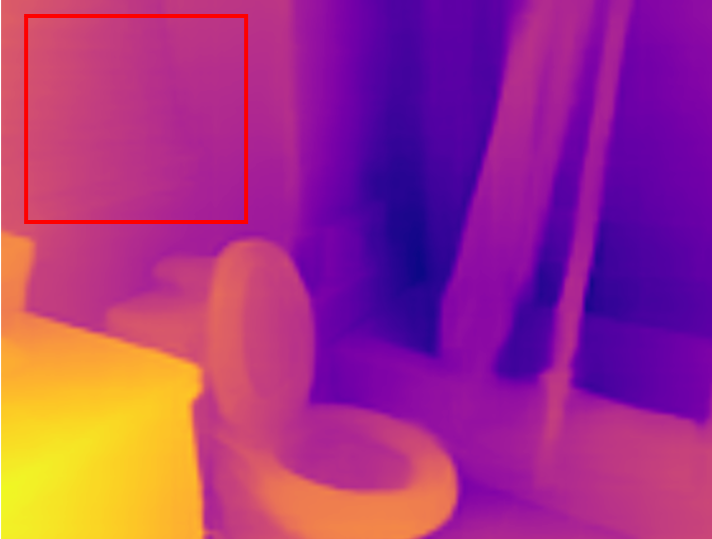}\\
        \end{minipage}
    }
    \subfigure[W/ Ours]{
        \begin{minipage}{0.16\linewidth}
        \centering
        \includegraphics[height=0.75in,width=1in]{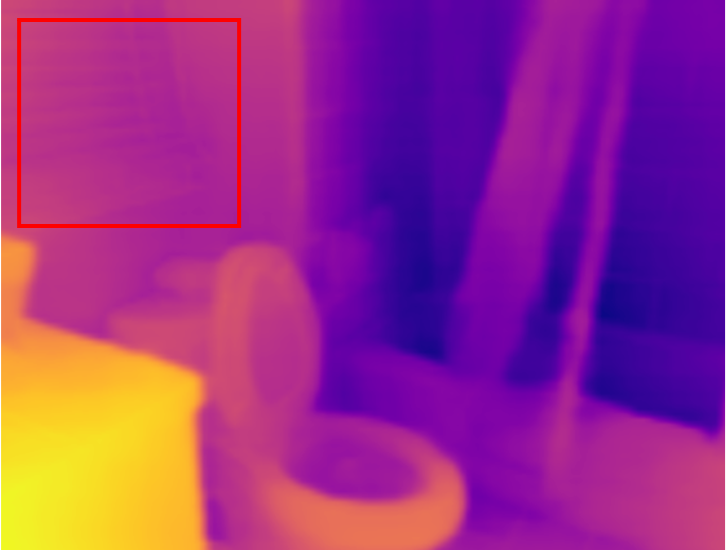}\\
        \end{minipage}
    }
    \subfigure[GT]{
        \begin{minipage}{0.16\linewidth}
        \centering
        \includegraphics[height=0.75in,width=1in]{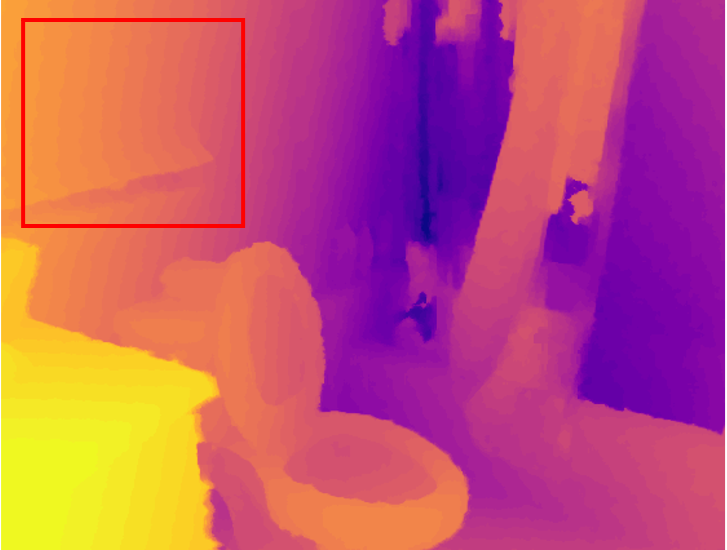}\\
        \end{minipage}
    }
    
    \caption{Visualization examples of predicted depth maps from selected MDE models on NYU Depth V2.}
    
    \label{fig:benchmark_nyu}
\end{figure}

\begin{figure*}[h]
\centering
    
    \rotatebox[origin=c]{90}{\scriptsize{LT}}
    \subfigure{
        \begin{minipage}{0.5\linewidth}
        \centering
        \includegraphics[height=0.8in]{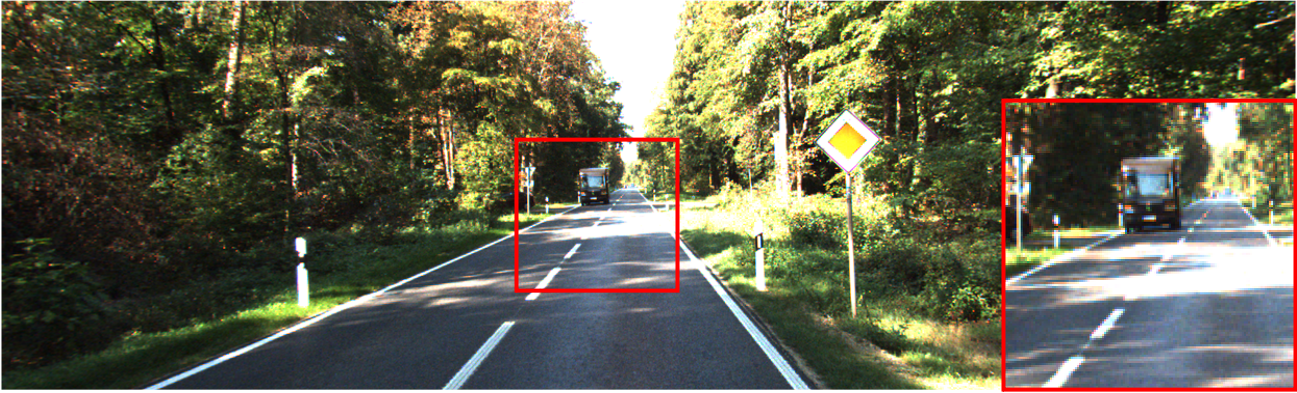} \\
        \end{minipage}
    }\hspace{-0.27in}
    \subfigure{
        \begin{minipage}{0.15\linewidth}
        \centering
        \includegraphics[height=0.8in,width=0.8in]{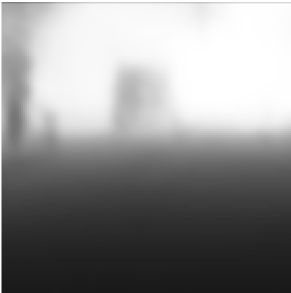} \\
        \end{minipage}
    }\hspace{-0.15in}
    \subfigure{
        \begin{minipage}{0.15\linewidth}
        \centering
        \includegraphics[height=0.8in,width=0.8in]{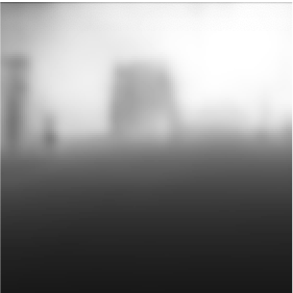} \\
        \end{minipage}
    }\hspace{-0.15in}
    \subfigure{
        \begin{minipage}{0.15\linewidth}
        \centering
        \includegraphics[height=0.8in,width=0.8in]{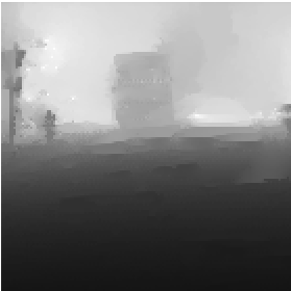} \\
        \end{minipage}
    }

    \vspace{-0.1in}
    \rotatebox[origin=c]{90}{\scriptsize{BTS}}
    \subfigure{
        \begin{minipage}{0.5\linewidth}
        \centering
        \includegraphics[height=0.8in]{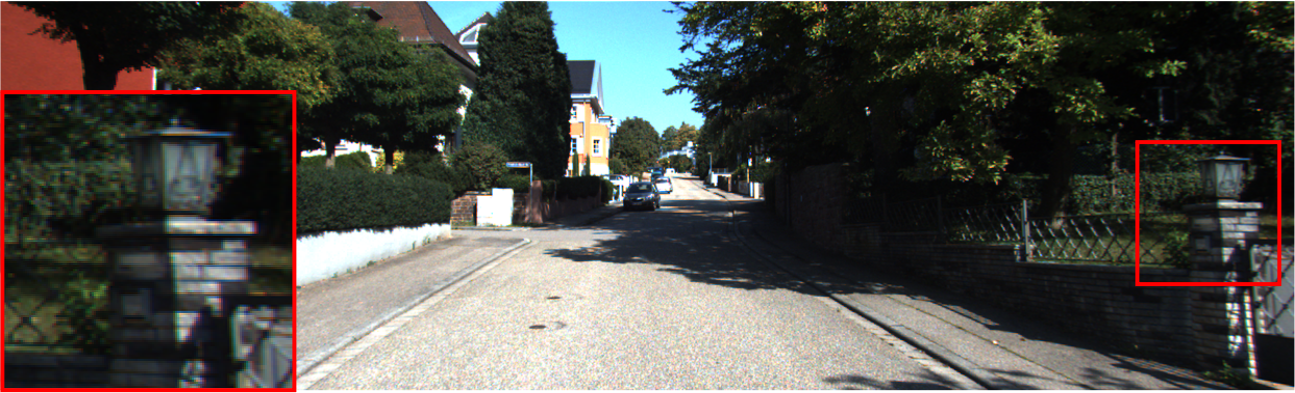} \\
        \end{minipage}
    }\hspace{-0.27in}
    \subfigure{
        \begin{minipage}{0.15\linewidth}
        \centering
        \includegraphics[height=0.8in,width=0.8in]{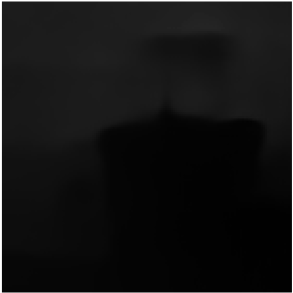} \\
        \end{minipage}
    }\hspace{-0.15in}
    \subfigure{
        \begin{minipage}{0.15\linewidth}
        \centering
        \includegraphics[height=0.8in,width=0.8in]{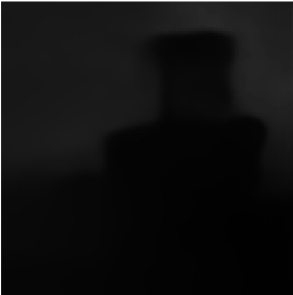} \\
        \end{minipage}
    }\hspace{-0.15in}
    \subfigure{
        \begin{minipage}{0.15\linewidth}
        \centering
        \includegraphics[height=0.8in,width=0.8in]{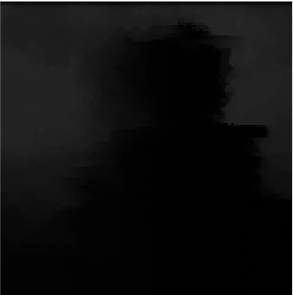} \\
        \end{minipage}
    }

    \setcounter{subfigure}{0}
    \vspace{-0.1in}
    \rotatebox[origin=c]{90}{\scriptsize{NeWCRFs}}
    \subfigure[RGB]{
        \begin{minipage}{0.5\linewidth}
        \centering
        \includegraphics[height=0.8in]{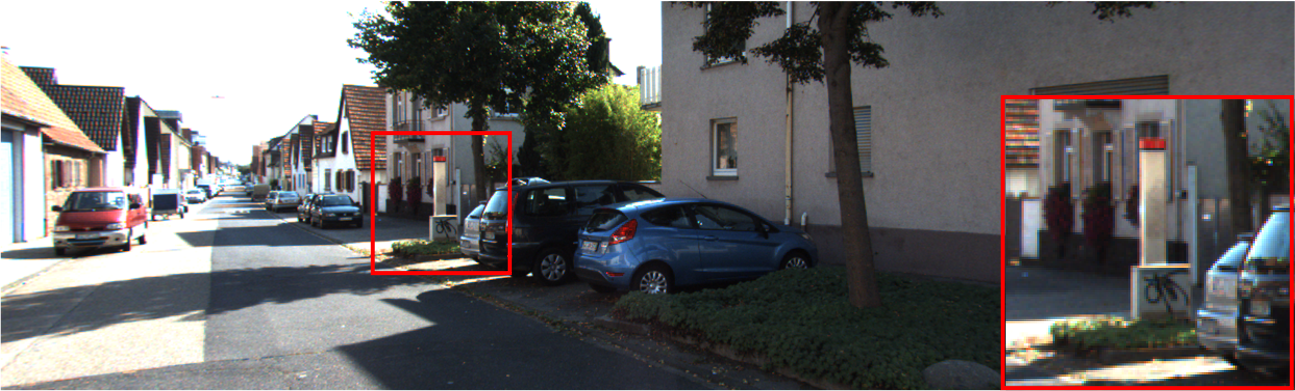}
        \end{minipage}
    }\hspace{-0.27in}
    \subfigure[Original]{
        \begin{minipage}{0.15\linewidth}
        \centering
        \includegraphics[height=0.8in,width=0.8in]{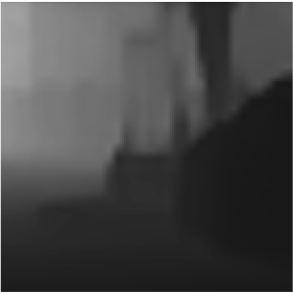}
        \end{minipage}
    }\hspace{-0.15in}
    \subfigure[W/ Ours]{
        \begin{minipage}{0.15\linewidth}
        \centering
        \includegraphics[height=0.8in,width=0.8in]{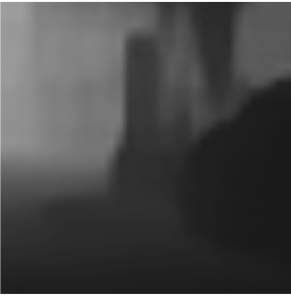}
        \end{minipage}
    }\hspace{-0.15in}
    \subfigure[GT]{
        \begin{minipage}{0.15\linewidth}
        \centering
        \includegraphics[height=0.8in,width=0.8in]{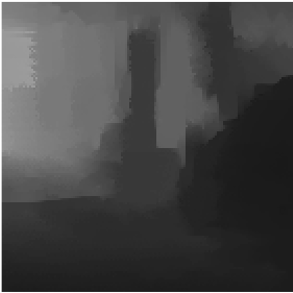}
        \end{minipage}
    }

    \caption{Visualization examples of predicted depth maps from selected MDE models on KITTI.}
    
    \label{fig:benchmark_kitti}
    
\end{figure*}

\begin{figure}[h]
    \centering
        
        \rotatebox[origin=c]{90}{\scriptsize{LT}}
        \subfigure{
            \begin{minipage}{0.16\linewidth}
            \centering
            \includegraphics[height=0.75in,width=1in]{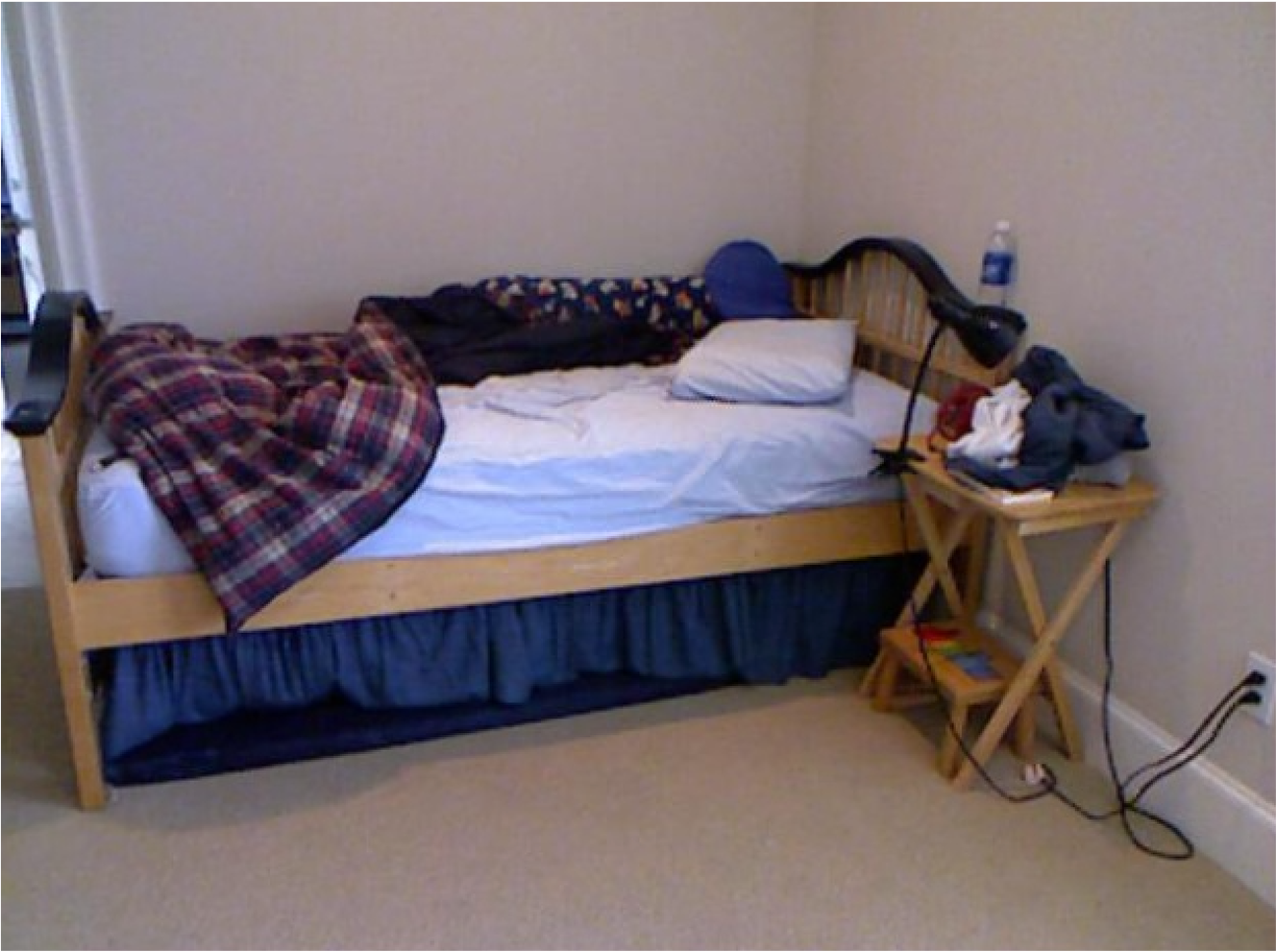}
            \end{minipage}
        }
        \subfigure{
            \begin{minipage}{0.16\linewidth}
            \centering
            \includegraphics[height=0.75in,width=1in]{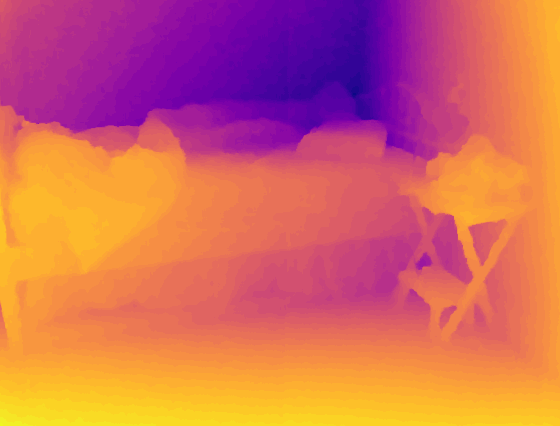}
            \end{minipage}
        }
        \subfigure{
            \begin{minipage}{0.16\linewidth}
            \centering
            \includegraphics[height=0.75in,width=1in]{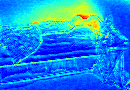}
            \end{minipage}
        }
        \subfigure{
            \begin{minipage}{0.16\linewidth}
            \centering
            \includegraphics[height=0.75in,width=1in]{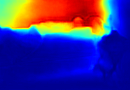}
            \end{minipage}
        }
        
        
        \vspace{-0.1in}
        \rotatebox[origin=c]{90}{\scriptsize{BTS}}
        \subfigure{
            \begin{minipage}{0.16\linewidth}
            \centering
            \includegraphics[height=0.75in,width=1in]{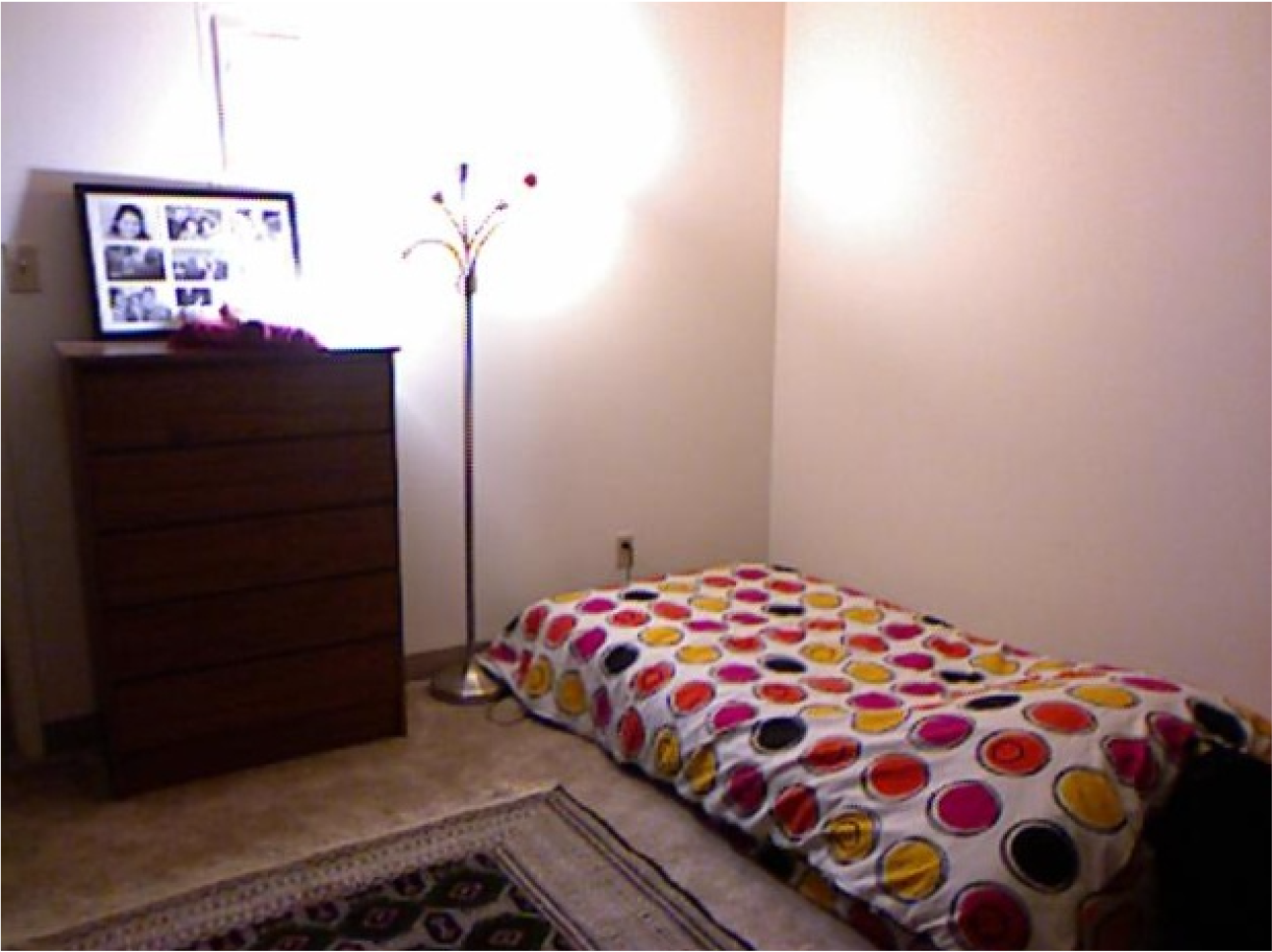}
            \end{minipage}
        }
        \subfigure{
            \begin{minipage}{0.16\linewidth}
            \centering
            \includegraphics[height=0.75in,width=1in]{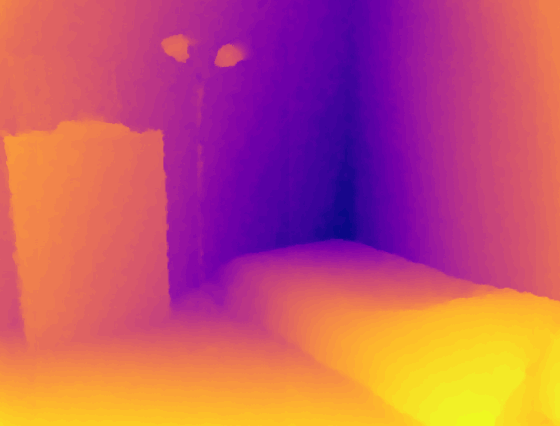}
            \end{minipage}
        }
        \subfigure{
            \begin{minipage}{0.16\linewidth}
            \centering
            \includegraphics[height=0.75in,width=1in]{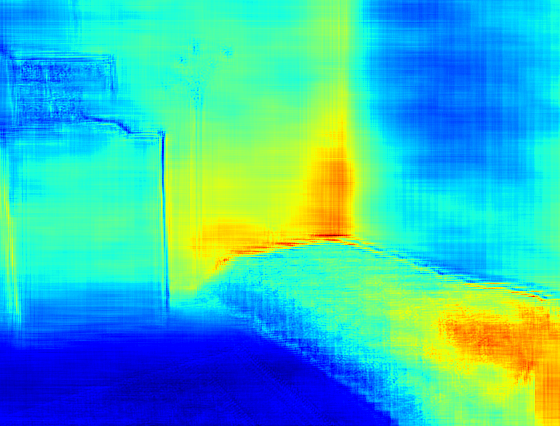}
            \end{minipage}
        }
        \subfigure{
            \begin{minipage}{0.16\linewidth}
            \centering
            \includegraphics[height=0.75in,width=1in]{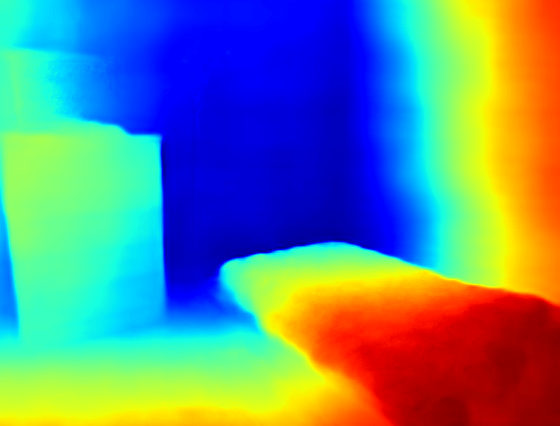}
            \end{minipage}
        }
        
        \vspace{-0.1in}
        \setcounter{subfigure}{0}
        
       \rotatebox[origin=c]{90}{\scriptsize{NeWCRFs}}
        \subfigure[RGB]{
            \begin{minipage}{0.16\linewidth}
            \centering
            \includegraphics[height=0.75in,width=1in]{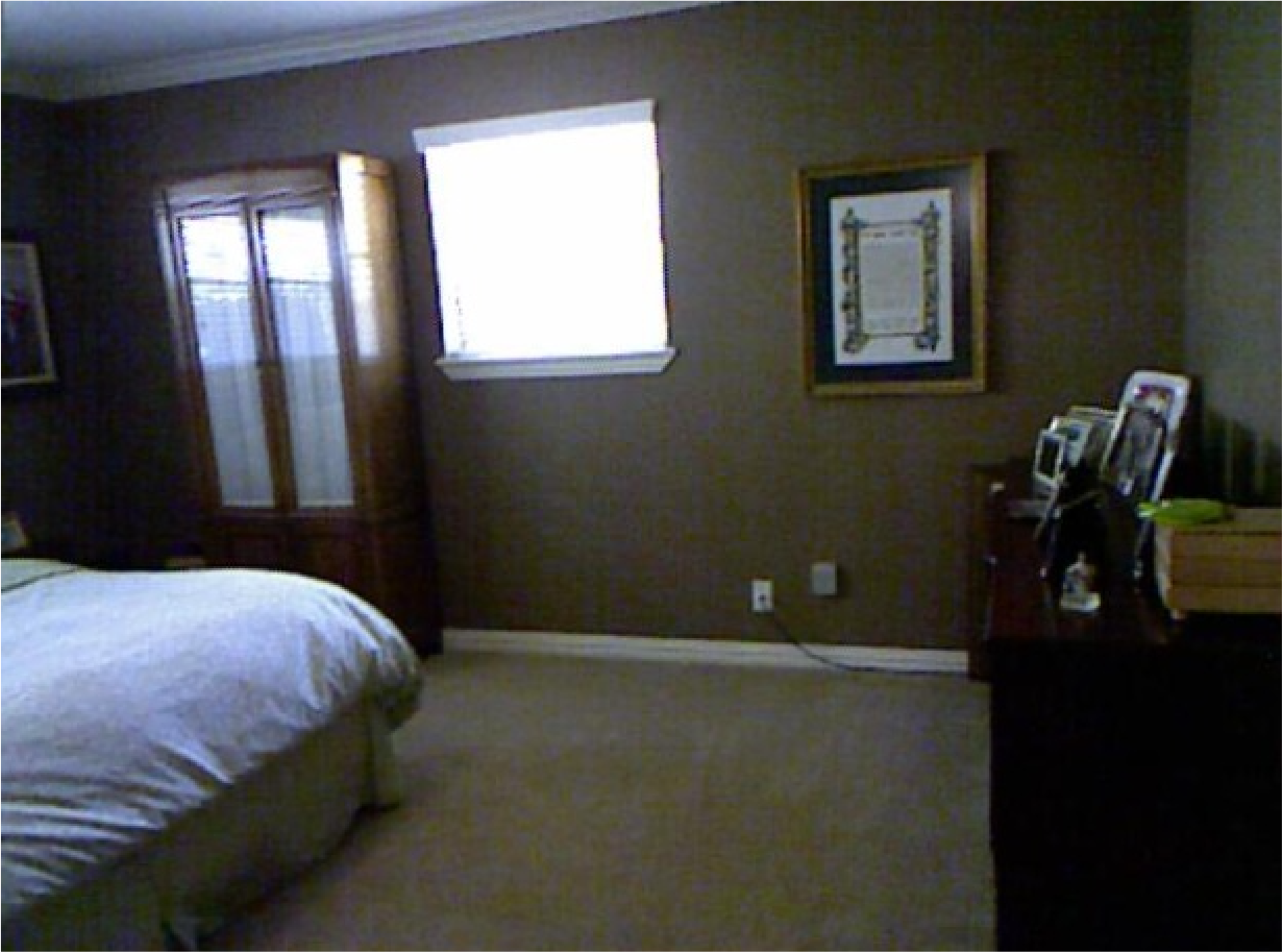}
            \end{minipage}
        }
        \subfigure[Depth]{
            \begin{minipage}{0.16\linewidth}
            \centering
            \includegraphics[height=0.75in,width=1in]{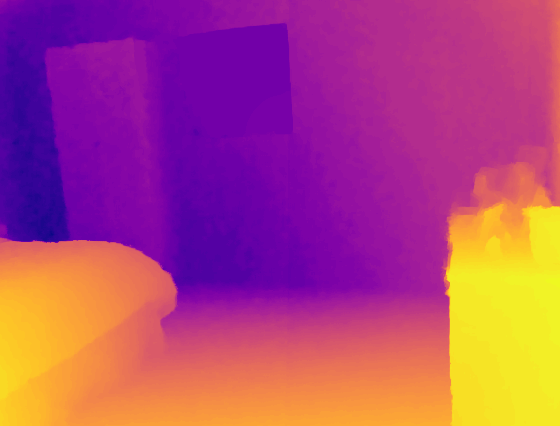}
            \end{minipage}
        }
        \subfigure[Original]{
            \begin{minipage}{0.16\linewidth}
            \centering
            \includegraphics[height=0.75in,width=1in]{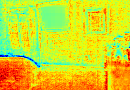}
            \end{minipage}
        }
        \subfigure[W/ Ours]{
            \begin{minipage}{0.16\linewidth}
            \centering
            \includegraphics[height=0.75in,width=1in]{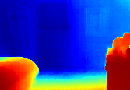}
            \end{minipage}
        }
        
        \caption{Visualizations on feature maps from selected models on NYU Depth V2.}
        
        \label{fig:feat_vis_nyu}
    \end{figure}

\begin{figure}[!h]
\centering
    
    \rotatebox[origin=c]{90}{\scriptsize{LT}}
    \subfigure{
        \begin{minipage}{0.16\linewidth}
        \centering
        \includegraphics[height=0.75in,width=1in]{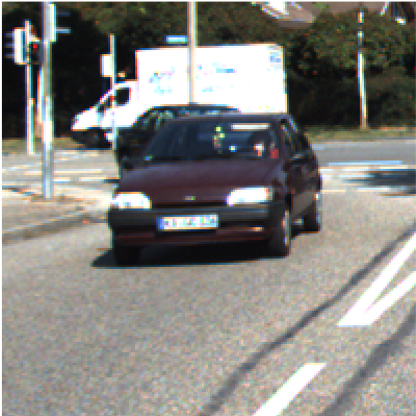}
        \end{minipage}
    }
    \subfigure{
        \begin{minipage}{0.16\linewidth}
        \centering
        \includegraphics[height=0.75in,width=1in]{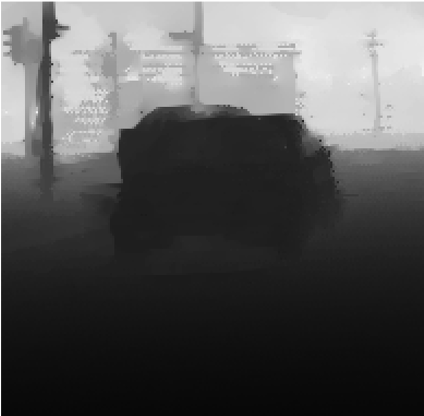}
        \end{minipage}
    }
    \subfigure{
        \begin{minipage}{0.16\linewidth}
        \centering
        \includegraphics[height=0.75in,width=1in]{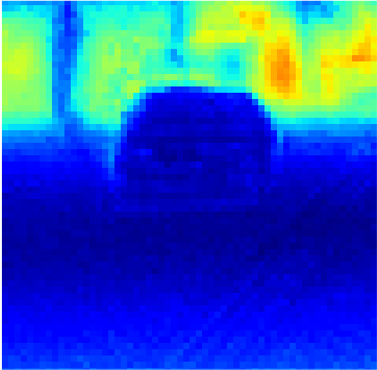}
        \end{minipage}
    }
    \subfigure{
        \begin{minipage}{0.16\linewidth}
        \centering
        \includegraphics[height=0.75in,width=1in]{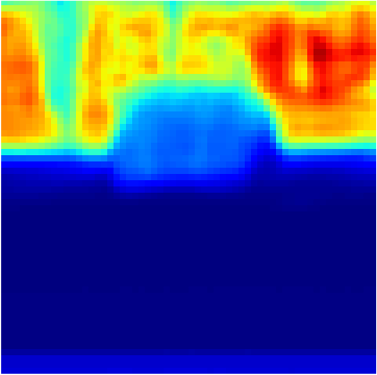}
        \end{minipage}
    }
    
    
    \vspace{-0.1in}
    \rotatebox[origin=c]{90}{\scriptsize{BTS}}
    \subfigure{
        \begin{minipage}{0.16\linewidth}
        \centering
        \includegraphics[height=0.75in,width=1in]{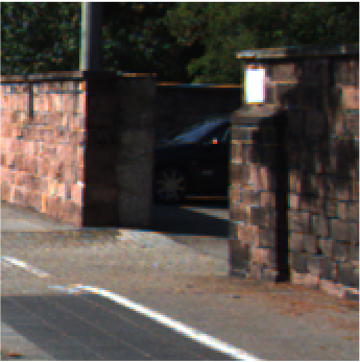}
        \end{minipage}
    }
    \subfigure{
        \begin{minipage}{0.16\linewidth}
        \centering
        \includegraphics[height=0.75in,width=1in]{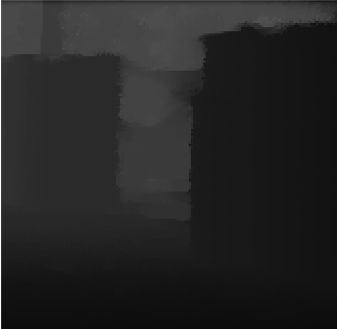}
        \end{minipage}
    }
    \subfigure{
        \begin{minipage}{0.16\linewidth}
        \centering
        \includegraphics[height=0.75in,width=1in]{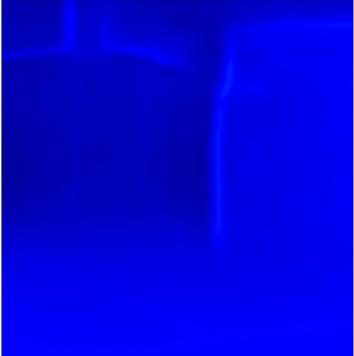}
        \end{minipage}
    }
    \subfigure{
        \begin{minipage}{0.16\linewidth}
        \centering
        \includegraphics[height=0.75in,width=1in]{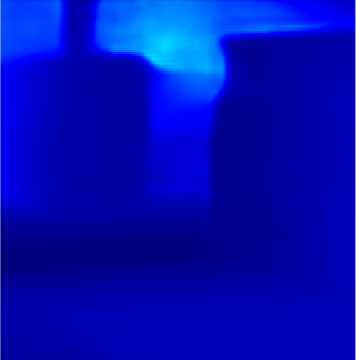}
        \end{minipage}
    }
    
    \vspace{-0.1in}
    \setcounter{subfigure}{0}
    
   \rotatebox[origin=c]{90}{\scriptsize{NeWCRFs}}
    \subfigure[RGB]{
        \begin{minipage}{0.16\linewidth}
        \centering
        \includegraphics[height=0.75in,width=1in]{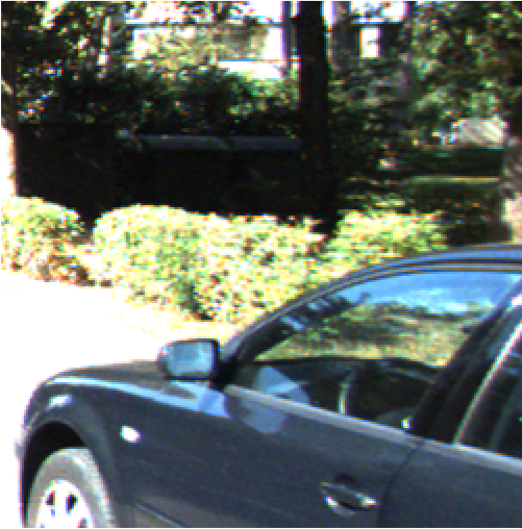}
        \end{minipage}
    }
    \subfigure[Depth]{
        \begin{minipage}{0.16\linewidth}
        \centering
        \includegraphics[height=0.75in,width=1in]{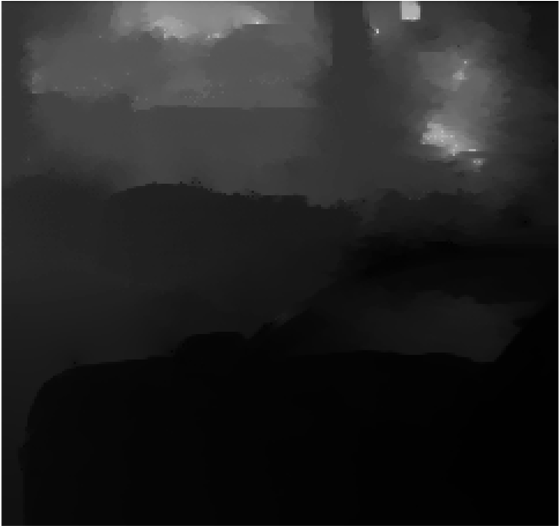}
        \end{minipage}
    }
    \subfigure[Original]{
        \begin{minipage}{0.16\linewidth}
        \centering
        \includegraphics[height=0.75in,width=1in]{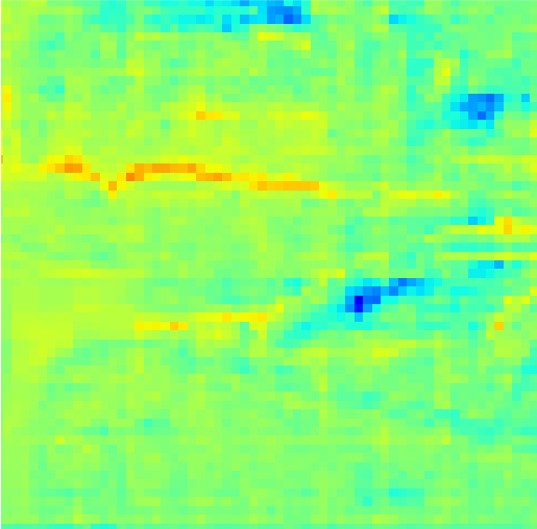}
        \end{minipage}
    }
    \subfigure[W/ Ours]{
        \begin{minipage}{0.16\linewidth}
        \centering
        \includegraphics[height=0.75in,width=1in]{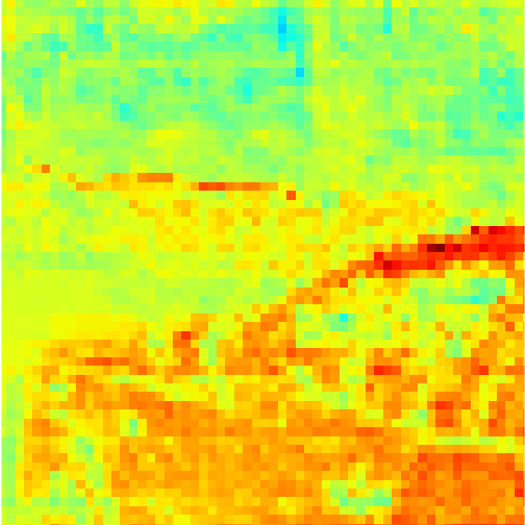}
        \end{minipage}
    }
    
    \caption{Visualizations on feature maps from selected models on KITTI.}
    
    \label{fig:feat_vis_kitti}
\end{figure}

The evaluation results from various MDE models across different datasets underscore the capacity of MetricDepth
 to enhance accuracy in a cross-model and cross-scene context.
 Beyond these quantitative results, we also present visualizations of the estimated depth maps and
 intermediate features from the selected MDE models to provide a more intuitive understanding of the impact of our method.
 Fig. \ref{fig:benchmark_nyu} and Fig. \ref{fig:benchmark_kitti} display the visualizations of the
 predicted depth maps on NYU Depth V2 and KITTI datasets.
 Observations from these examples reveal that, with the integration of our method during model training,
 the MDE models exhibit an improved ability to perceive depth information in scenarios involving thin objects
 and objects with complex details.
 This enhancement is consistent across both outdoor and indoor scenes, indicating the capability of our method
 in improving overall accuracy and detail perception.
 Further, Fig. \ref{fig:feat_vis_nyu} and Fig. \ref{fig:feat_vis_kitti} offer visualizations of the deep features
 from selected MDE models with and without applying our method.
 These comparisons highlight that deep features regularized by our method align more closely with the
 distributions of depth values in ground truth annotations.
 Besides, more discriminating deep features are presented in the areas experiencing sharp depth transitions.
 These visual demonstrations of the deep features affirm that our method effectively regularizes the feature space,
 leading to improved quality in depth prediction.
 Moreover, these visualizations validate that by exploring the characteristics of deep features in MDE models, the
 quality of predicted depth can be truly improved.

\subsection{Ablation Studies}
\label{sec:exp_ab}

To inspect the influence of different components in MetricDepth on the final prediction performance, numerous
 experiments are conducted. In this section, the evaluation results of these experiments are presented.

\subsubsection{Effect of Feature Sample Collecting}
In MetricDepth, two means of feature sample collecting are adopted, which are ``within current feature map''
 and ``across the training batch''. 
 In this section, the experimental results demonstrating the proper sample count for these two
 sample collecting means and the effect of them on prediction performance are shown.

\begin{figure}[t]
    \centering
    \subfigure{
    \includegraphics[width=0.45\textwidth]{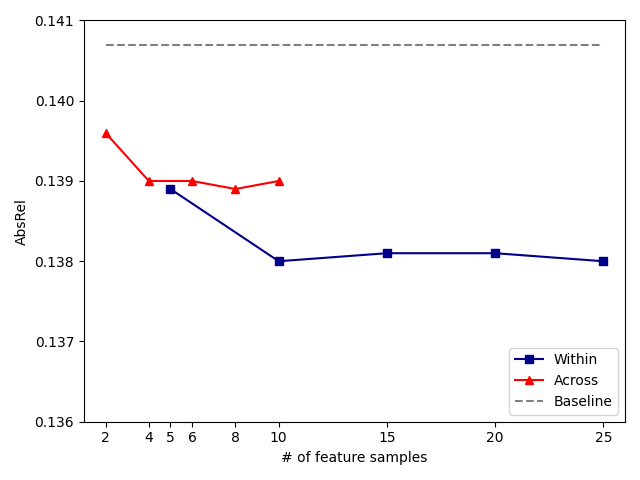}
    }
    \caption{The change of $AbsRel$  with different number of feature samples under two kinds of feature sample
    collecting means.
    }
    \label{fig:sample_count}
\end{figure}

Fig. \ref{fig:sample_count} illustrates the $AbsRel$ (lower is better) evaluation index 
 for different quantities of feature samples collected exclusively using these two means.
 As indicated in Fig. \ref{fig:sample_count}, for the feature sample collecting means
 ``within current feature map'', gathering 10 samples for anchor feature can lead to satisfied
 promotion effect on baseline model.
 When more feature samples are involved, the prediction performance hardly get further promoted.
 The situation is similar for the collecting means ``across the training batch''.
 Seen from Fig. \ref{fig:sample_count},
 collecting 4 sample feature maps across the training batch for anchor feature map is the suitable choice
 to achieve acceptable estimation performance.
 More detailed experiment data is shown in Table \ref{tab:count_within} and Table \ref{tab:count_across}.

\begin{table}[h]

    \begin{minipage}{0.45\textwidth}
    \caption{Evaluation results with different feature sample counts under the ``within current feature map''
        feature sample collecting means.}
        \centering
        \begin{tabular}{cccc}
        \toprule
        Sample Count & $AbsRel\downarrow $  & $RMSE\downarrow$ & $\delta < 1.25\uparrow $ \\
        \midrule
        5  & 0.1389 & 0.4756 & 0.8118  \\
        10 & 0.1380 & 0.4745 & 0.8131  \\
        15 & 0.1381 & 0.4744 & 0.8133 \\
        20 & 0.1381 & 0.4747 & 0.8130  \\
        25 & 0.1380 & 0.4742 & 0.8134  \\
        \bottomrule
    \end{tabular}
    \label{tab:count_within}
    \end{minipage}
    \begin{minipage}{0.45\textwidth}
    \caption{Evaluation results with different feature sample counts under the ``across the training batch''
    feature sample collecting means.}
    \centering
    \begin{tabular}{cccc}
    \toprule
      Sample Count & $AbsRel\downarrow$  & $RMSE\downarrow$ & $\delta < 1.25\uparrow $  \\
      \midrule
      2  & 0.1396 & 0.4812 & 0.8083 \\
      4  & 0.1390 & 0.4760 & 0.8119 \\
      6  & 0.1390 & 0.4762 & 0.8121 \\
      8  & 0.1389 & 0.4759 & 0.8122 \\
      10 & 0.1390 & 0.4763 & 0.8122 \\
      \bottomrule
    \end{tabular}
    \label{tab:count_across}
    \end{minipage}
\end{table}

    

Table \ref{tab:collecting_both} compares the results when these two sample collection means are applied either
 independently or in conjunction.
 Even a single sample collecting means can provide valid feature samples for feature regularizing, the incorporation
 of both sample collecting means can collect more comprehensive feature samples from diverse scene locations,
 which additionally boosts the performance of feature regularizing process.

\begin{table}[h]
    \caption{Evaluation results with the combination of both feature sample collecting means.}
    \centering
    \begin{tabular}{ccccc}
    \toprule
      Collecting Type & Sample Count   & $AbsRel\downarrow$  & $RMSE\downarrow$ & $\delta < 1.25\uparrow $  \\
      \midrule
      Within          & 10 & 0.1380 & 0.4745 & 0.8131 \\
      Across          & 4  & 0.1390 & 0.4760 & 0.8119 \\
      \textbf{Within and Across}   & \textbf{10 and 4} & \textbf{0.1375} & \textbf{0.4734} & \textbf{0.8142} \\
      \bottomrule
    \end{tabular}
    \label{tab:collecting_both}
\end{table}

Above experiments exploring the effect of different feature sample collecting means are conducted
 under the condition that $r_{p} = 0.1$, $r_{l}=0.5$, $r_{h}=1$, and $m_{ra}=2$.
 Taking into consideration that 10 and 4 feature samples respectively for the sample collecting means
 ``within current feature map'' and ``across the training batch'' can lead to satisfied performance,
 the experiments in other sections obey this configuration unless specially announced.

\subsubsection{Configuration of the Multi-Range Strategy}

The main motivation of the multi-range strategy is refining the identification case on negative samples according to
 diverse ranges of depth differential to achieve differentiated regularizing between anchor and negative samples.
 In this section, the experimental results with different
 identification cases and diverse regularizing margins on negative feature samples are reported.
 In these experiments, negative samples are identified into different subgroups according to depth differential range
 with interval of 0.5.
 The evaluation results employing multiple subgroups of negative samples and different regularizing margins
 are shown in Table \ref{tab:range_grouping}.

\begin{table}[h]
    \caption{Evaluation results under the multi-range strategy with different identification cases and diverse regularizing
    margins on negative samples.}
    \centering
    \begin{tabular}{cccccc}
    \toprule
     $r_{p}$  & $r_{l}$ - $r_{h}$  &  $m_{ra}$  & $AbsRel\downarrow$  & $RMSE\downarrow$ & $\delta < 1.25\uparrow $  \\
      \midrule
      -   &-                 & -             & 0.1407 & 0.4862 & 0.8058 \\
      \midrule
      0.1   &0.5-1                 & 2             & 0.1377 & 0.4754 & 0.8152 \\
      \textbf{0.1}   &\textbf{0.5-1}        & \textbf{3}    & \textbf{0.1372} & \textbf{0.4749} & \textbf{0.8161} \\
      0.1   &0.5-1                 & 4             & 0.1376 & 0.4764 & 0.8117 \\
      0.1   &0.5-1                 & 5             & 0.1379 & 0.4870 & 0.8098 \\
      \midrule
      0.1   &0.5-1, 1-1.5          & 3, 4          & 0.1373 & 0.4781 & 0.8140 \\
      0.1   &0.5-1, 1-1.5          & 3, 5          & 0.1371 & 0.4750 & 0.8169 \\
      \textbf{0.1}   &\textbf{0.5-1, 1-1.5} & \textbf{3, 6} & \textbf{0.1368} & \textbf{0.4746} & \textbf{0.8180} \\
      0.1   &0.5-1, 1-1.5          & 3, 7          & 0.1377 & 0.4750 & 0.8153 \\
      \midrule
      
      0.1   &0.5-1, 1-1.5, 1.5-2   & 3, 6 ,7       & 0.1372 & 0.4740 & 0.8163 \\
      \textbf{0.1}   &\textbf{0.5-1, 1-1.5, 1.5-2}   & \textbf{3, 6, 8}       & \textbf{0.1365} & \textbf{0.4731} & \textbf{0.8182} \\
      0.1   &0.5-1, 1-1.5, 1.5-2   & 3, 6, 9       & 0.1373 & 0.4767 & 0.8177 \\
      0.1   &0.5-1, 1-1.5, 1.5-2   & 3, 6, 10      & 0.1384 & 0.4802 & 0.8142 \\
      \bottomrule
    \end{tabular}
    \label{tab:range_grouping}
\end{table}


An analysis of the data in Table \ref{tab:range_grouping} reveals that the choice of regularizing margin for
 negative samples within specific depth differential range significantly influences the final prediction performance.
 Notably, the application of an inappropriate regularizing margin on negative samples in a certain range of depth differential
 can negatively impact the regularization effectiveness, thereby diminishing the accuracy enhancement potential of our method.
 For example, applying an excessively large regularizing margin ($m_{ra}=5$) to negative samples
 in the 0.5-1m depth differential range leads to inferior performance compared to a more optimal margin ($m_{ra}=2$).
 For negative samples with larger depth differentials relative to the anchor,
 a higher regularizing margin is more suitable.
 Moreover, by adjusting appropriate regularizing margins for negative samples across multiple depth differential ranges,
 the performance of our method is further enhanced.
 The analysis from the data in Table \ref{tab:range_grouping} also validates the rationale behind our multi-range strategy:
 for negative samples with diverse depth differentials to anchor,
 the identification cases and regularizing manners should be differentiated.

 
Additionally, this section compares the performance of applying the uniform strategy versus the multi-range strategy in MetricDepth.
 Table \ref{tab:binary_grouping} presents the evaluation results under the uniform strategy,
 considering negative samples from various depth differential ranges and different implementations on regularizing margins.
 Table \ref{tab:grouping_cmp} showcases the comparison of peak performance achieved by
 both the uniform strategy and the multi-range strategy.
 While the application of the uniform strategy in MetricDepth can also improve the prediction accuracy of baseline model,
 the integration of the multi-range strategy results in even superior performance.
 This comparison not only underscores the effectiveness of the multi-range strategy
 but also highlights its advantages and rationality.

\begin{table}[h]
    \caption{Evaluation results under the uniform strategy with different identification cases and
    diverse regularizing margins on negative samples.}
    \centering
    \begin{tabular}{cccccc}
    \toprule
     $r_{p}$ & $r_{n}$           &  $m_{u}$          & $AbsRel\downarrow$  & $RMSE\downarrow$ & $\delta < 1.25\uparrow $  \\
      \midrule
      -   &-                 & -             & 0.1407 & 0.4862 & 0.8058 \\
      \midrule
      0.1  & 0.5               & 3             & 0.1377 & 0.4755 & 0.8151 \\
      \textbf{0.1}  & \textbf{0.5}   & \textbf{4}             & \textbf{0.1371} & \textbf{0.4745} & \textbf{0.8159} \\
      0.1  & 0.5               & 5             & 0.1379 & 0.4780 & 0.8149 \\
      0.1  & 0.5               & 6             & 0.1382 & 0.4789 & 0.8142 \\
      \midrule
      0.1  & 1                 & 3             & 0.1383 & 0.4804 & 0.8142 \\
      0.1  & 1                 & 4             & 0.1379 & 0.4776 & 0.8149 \\
      \textbf{0.1}  & \textbf{1}   & \textbf{5}   & \textbf{0.1376} & \textbf{0.4756} & \textbf{0.8156} \\
      0.1  & 1                 & 6             & 0.1385 & 0.4811 & 0.8139 \\
      \midrule
      0.1  & 1.5               & 3             & 0.1396 & 0.4831 & 0.8119 \\
      0.1  & 1.5               & 4             & 0.1392 & 0.4823 & 0.8135 \\
      \textbf{0.1} & \textbf{1.5} & \textbf{5} & \textbf{0.1387} & \textbf{0.4771} & \textbf{0.8148} \\
      0.1  & 1.5               & 6             & 0.1395 & 0.4829 & 0.8128 \\
      \bottomrule
    \end{tabular}
    \label{tab:binary_grouping}
\end{table}

\begin{table}[h]
    \caption{Performance comparison between the uniform strategy and the multi-range strategy.}
    \centering
    \begin{tabular}{cccc}
    
    \toprule
    Type          & $AbsRel\downarrow$  & $RMSE\downarrow$ & $\delta < 1.25\uparrow $  \\
    \midrule
    Baseline        & 0.1407 & 0.4862 & 0.8058 \\
    Uniform   & 0.1371 & 0.4745 & 0.8159 \\
    \textbf{Multi-Range}    & \textbf{0.1365} & \textbf{0.4731} & \textbf{0.8182} \\
    \bottomrule
    
    \end{tabular}
    \label{tab:grouping_cmp}
\end{table}
\section{Conclusion}

In this work, we introduce MetricDepth, an approach which leverages deep metric learning to enhance the estimation performance
 for MDE task.
 By employing the novel differential-based sample identification, MetricDepth innovatively distinguishes feature samples
 as different types according to their depth differential to anchor, enabling the sample type identification in MDE task.
 In addition, we propose the multi-range strategy to better handle negative samples having diverse depth differentials
 to anchor, strengthening the regularizing capability of MetricDepth.
 The comprehensive experiments on widely-used MDE datasets showcase that substantial improvements across
 different models of various architectures can be achieved with our method,
 validating the effectiveness and versatility of our method.
 These experimental results also reveal that mining the potentials of deep features in MDE models is a promising way
 in advancing the performance for MDE task.

Despite encouraging results has been achieved with MetricDepth, we acknowledge that there also exist limitations in our method.
 The current reliance on the hyperparameters of depth differential threshold for identification cases
 and regularizing margins on negative samples presents challenges in finding the optimal choice.
 In future work, we will try to develop more adaptive method to determine the identification case
 and regularization margins on negative sample autonomously, thus mitigating the need for manual hyperparameter tuning.

\bibliographystyle{IEEEtran}
\bibliography{ref}

\end{document}